%% file: M2Net.tex
\documentclass[conference]{IEEEtran}
\IEEEoverridecommandlockouts

\usepackage{cite}
\usepackage{amsmath,amssymb,amsfonts}
\usepackage{algorithm}
\usepackage{algpseudocode}
\usepackage{graphicx}
\usepackage{textcomp}
\usepackage[table]{xcolor}
\usepackage{booktabs}
\usepackage{multirow}
\usepackage{array}
\usepackage{subcaption}
\usepackage{enumitem}
\usepackage{tikz}
\usepackage{color, xspace}
\usepackage{pifont}
\usepackage{url}
\usepackage{makecell}

\setlist[itemize]{leftmargin=*, nosep}

\newcommand{\name}{{$M^2Net$}\xspace}

\newcommand{\eg}{\emph{e.g.,}\xspace}
\newcommand{\ie}{\emph{i.e.,}\xspace}

\def\BibTeX{{\rm B\kern-.05em{\sc i\kern-.025em b}\kern-.08em
    T\kern-.1667em\lower.7ex\hbox{E}\kern-.125emX}}
\begin{document}

\title{$M^2Net$: Unifying Scale-Specific Structures and Prototype Decomposition for Spatial-Temporal Forecasting}

\author{
\begin{tabular}{c}
Chengyang Zhou$^{1, 2, 3}$, Zijian Zhang$^{1, 3,\dagger}$, Hao Miao$^{4, \dagger}$, Chunxu Zhang$^{1, 3}$,\\
Ziming Liu$^{1}$, Chao Tang$^{1}$, Juncheng Hu$^{1,\dagger}$, Irwin King$^{5}$\\[0.25em]
\small
$^{1}$College of Computer Science and Technology, Jilin University, Changchun, China \\
\small
$^{2}$Institute of Computing Technology, Chinese Academy of Sciences, Beijing, China  \\
\small
$^{3}$Key Laboratory of Symbolic Computation and Knowledge Engineering of Ministry of Education, Jilin University, Changchun, China \\
\small
$^{4}$Shenzhen Institute for Advanced Study, University of Electronic Science and Technology of China, Shenzhen, China \\
\small
$^{5}$The Chinese University of Hong Kong, Hong Kong, China\\
\small
\{chengyang25, liuzm5523, tangzhao5523\}@mails.jlu.edu.cn,\quad
\{zhangzijian, zhangchunxu, jchu\}@jlu.edu.cn\\
\small
miaohao@uestc.edu.cn,\quad king@cse.cuhk.edu.hk
\end{tabular}
\thanks{$^{\dagger}$ Corresponding author: Zijian Zhang, Hao Miao and Juncheng Hu}
}

\maketitle

\begin{abstract}
Spatial-temporal forecasting is pivotal for intelligent urban management yet remains challenging due to the intricate coupling of spatial dynamics and temporal non-stationarity. 
Existing approaches, while effective, often overlook two fundamental characteristics of real-world data: 
(1) Granularity-dependence, where dominant spatial correlations shift significantly across different time scales, and 
(2) Signal entanglement, where global common patterns are inextricably mixed with node-specific disturbances. 
In this paper, we propose a novel framework named $M^2Net$ (Multi-Scale Masked attention and Prototype network) to unify scale-specific structural learning and semantic decomposition.
First, to capture the evolution of spatial dependencies, we introduce a Multi-Scale Masked attention mechanism. By downsampling temporal features into multiple resolutions, we dynamically generate scale-adaptive spatial masks, allowing the model to distinguish between micro-scale physical neighbors and macro-scale functional correlations.
Second, to disentangle unique variations from shared trends, we propose a Prototype Residual Learning module. It retrieves global common patterns from a momentum-updated prototype bank and subtracts them from node representations, explicitly forcing the network to model the residual unique deviations.
By integrating these two orthogonal views, i.e., structural multiscaling and semantic decomposition, our approach effectively isolates valid spatial-temporal signals from noise. Extensive experiments on real-world datasets demonstrate that $M^2Net$ outperforms state-of-the-art baselines on extensive empirical verifications.
\end{abstract}


\input{sections/2_Introduce}
\input{sections/4_Method}

\input{sections/5_Experiments}

\input{sections/3_RelatedWorks}
\input{sections/6_Conclusion}

\bibliographystyle{IEEEtran}
\bibliography{sections/8_References}

\end{document}

%% file: sections/2_Introduce.tex
\section{Introduction}

\begin{figure}[!t]
    \centering
    \begin{minipage}[t]{0.42\columnwidth}
        \vspace{0pt}
        \centering
        \includegraphics[width=\linewidth]{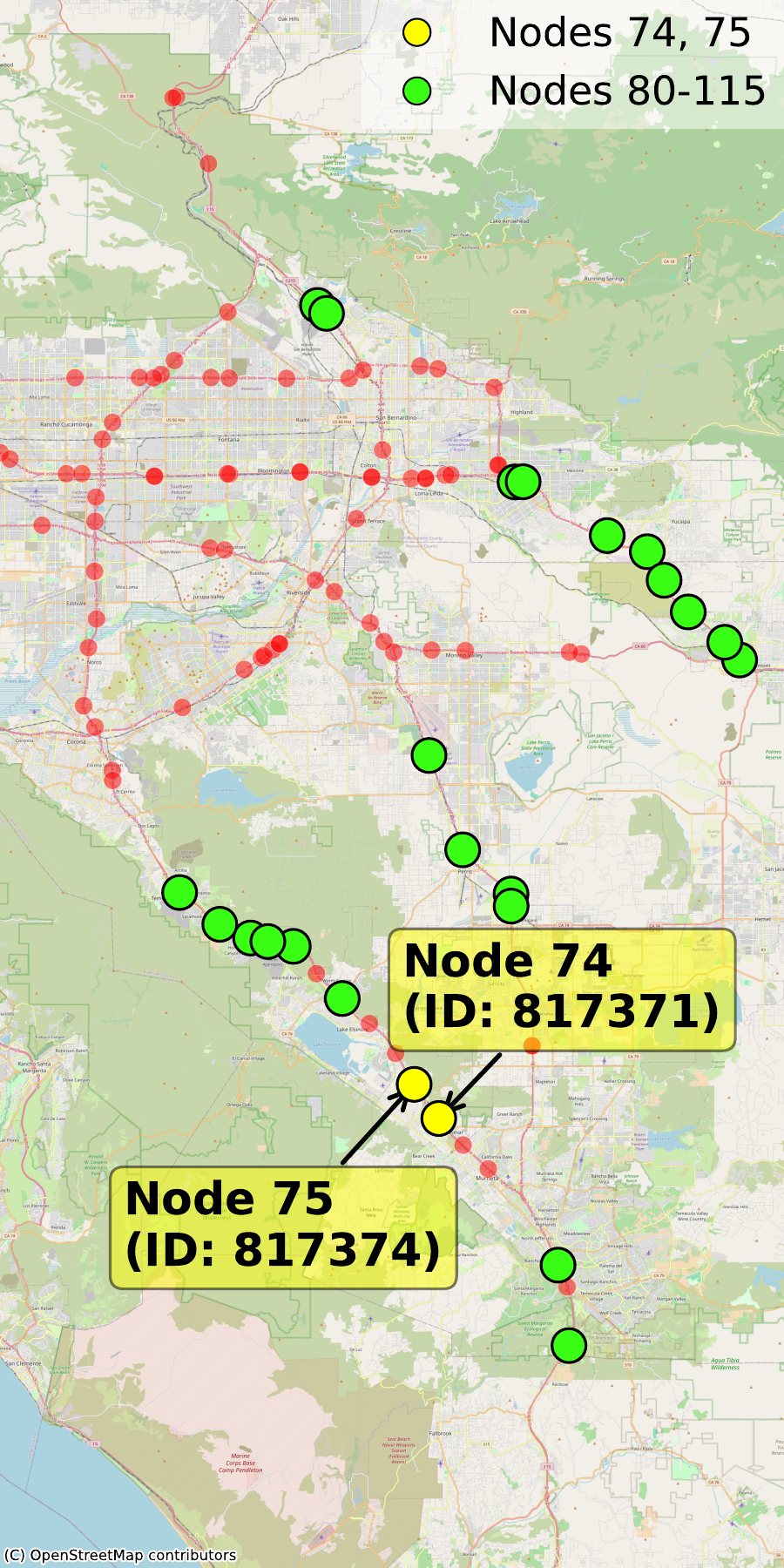}
        
        \vspace{14.5pt}
        \caption*{(a) Node distribution.}
    \end{minipage}
    \hfill
    \begin{minipage}[t]{0.54\columnwidth}
        \vspace{-8pt}
        \centering
        \includegraphics[width=\linewidth]{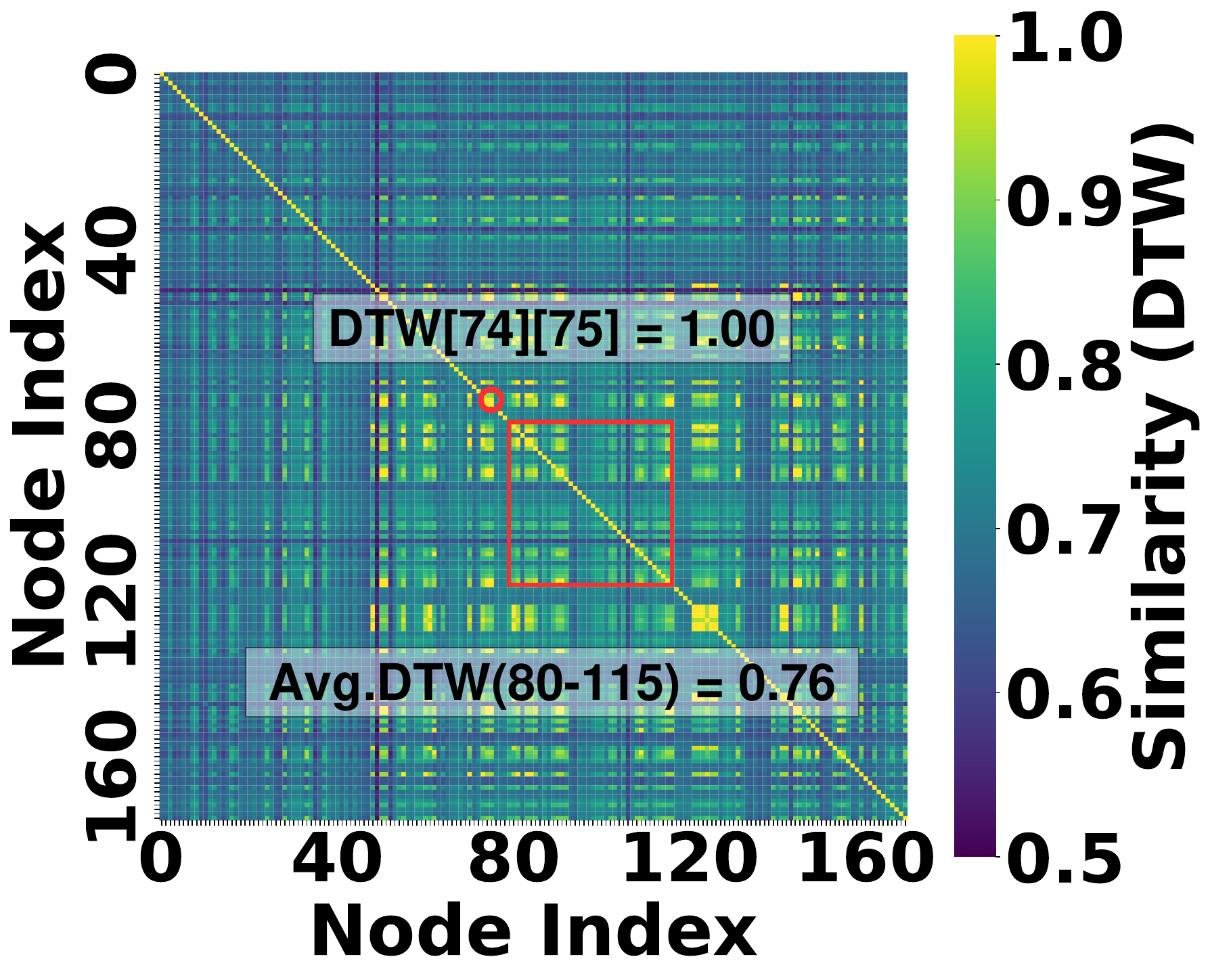}
        \vspace{-15pt}
        \caption*{(b) Node similarity (5-min).}

        \vspace{2pt}

        \includegraphics[width=\linewidth]{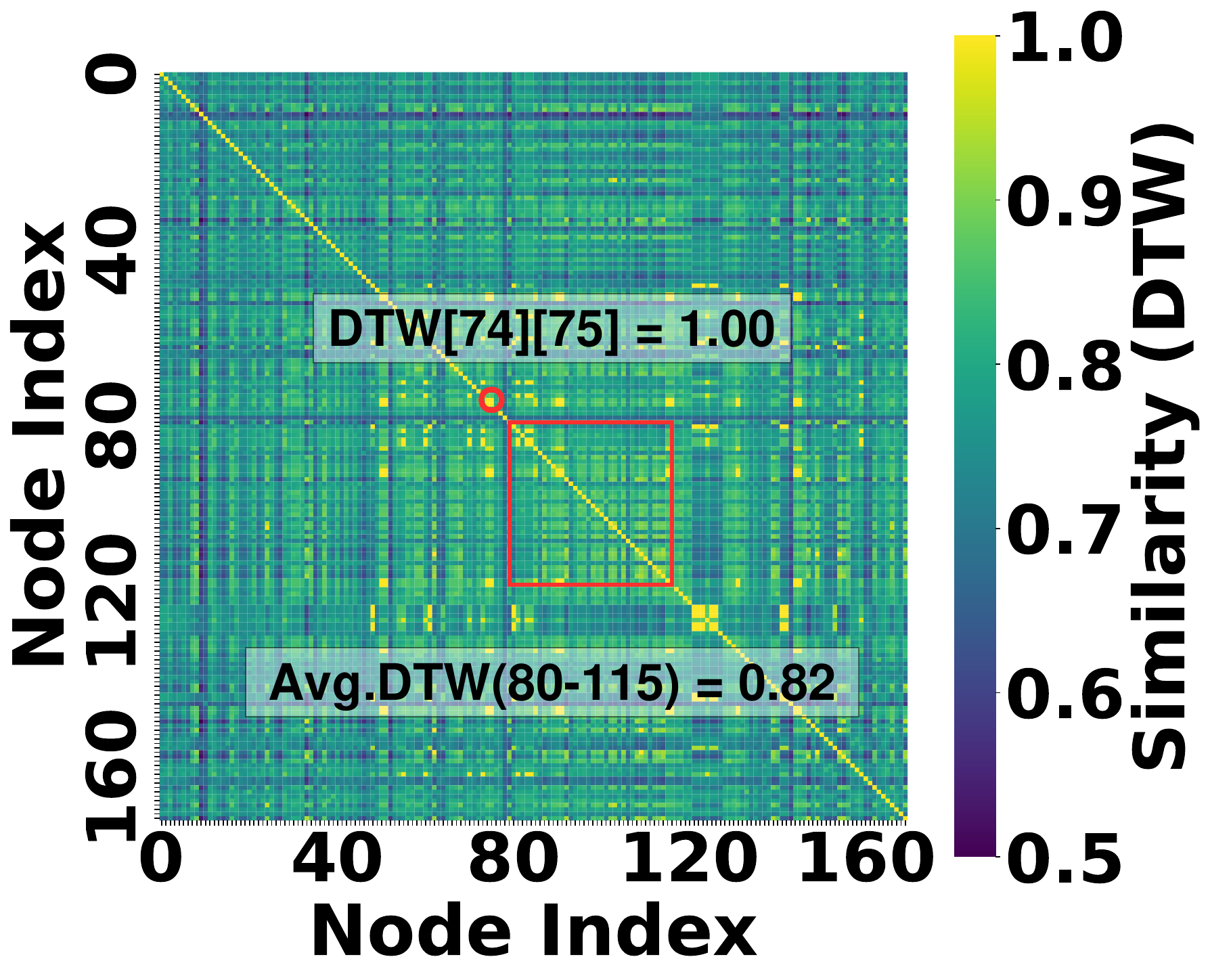}
        \vspace{-15pt}
        \caption*{(c) Node similarity (40-min).}
    \end{minipage}
    \caption{Spatial relation visualization of PEMS08. Adjacent nodes (\eg 74 and 75) exhibit high spatial correlation at both scales. Coarse temporal scale (c) captures more long-range, non-adjacent relationships (\eg nodes 80--115) compared to fine-grained scale (b).}
    \label{fig:half_column_analysis}
    \vspace{-6mm}
\end{figure}

The rapid deployment of sensing infrastructures and Internet-of-Things (IoT) devices has generated massive volumes of traffic data in modern cities~\cite{miao2024less, qiu2024tfb}. Effectively storing, managing, and analyzing these continuously collected spatial-temporal observations is critical for a variety of urban applications, including traffic management~\cite{deng2022multi, mo2020towards, yuan2024nuhuo, zhou2026feddis}, congestion mitigation~\cite{guo2015ecomark, tran2020deeptrans}, and route planning~\cite{xu2024managing, guo2024efficient, fang2021mdtp, shao2025hyperd}. 
As a result, modern spatial-temporal database systems and urban data management platforms have become essential infrastructures for handling large-scale traffic data~\cite{sakurai2015mining}. Among the core functionalities supported by these database systems, traffic flow forecasting~\cite{miao2022mba, tran2020deeptrans, xia2025self, jin2023spatio, zou2026meta}, which predicts future traffic conditions based on historical time series~\cite{chang2024timedrl, deng2022multi}, serves as a fundamental data management task, enabling proactive decision making and intelligent transportation services~\cite{cirstea2021enhancenet, dai2023dynamic}. 
This task is challenging because traffic dynamics are governed by a complex interplay of factors, including spatial correlations embedded in road networks and non-stationary temporal evolution of traffic states~\cite{capone2025spatio,guo2019attention}. 
Accurately capturing these dual dependencies is critical for database-driven analytics, enabling decision-makers to optimize resource allocation and deploy proactive congestion control mechanisms.

The evolution of spatial-temporal forecasting has been characterized by the pursuit of more expressive representations for these dual dependencies.
Early approaches, \eg GWNet~\cite{wu2019graph} and AGCRN~\cite{bai2020adaptive}, established a unified foundation by integrating adaptive graph learning with Recurrent Neural Networks. Recently, hybrid architectures, \eg STAEformer~{\cite{liu2023spatio}}, PDFormer~\cite{jiang2023pdformer}, and D2STGNN~\cite{shao2022decoupled}, have pushed performance boundaries by employing Transformer-based attention mechanisms or disentangled representation learning. 

Despite the promising performance of these advanced models, they still rely on spatial dependencies at a single temporal resolution, \ie either through a fixed graph structure or an adaptive matrix applied to the entire time series. 
This design implicitly assumes that spatial correlations remain constant regardless of temporal granularity, overlooking a critical characteristic: the dominant spatial dependencies shift significantly across different time scales.
 
\textbf{However, in reality, the effective spatial structure is inherently dependent on temporal granularity.} 
As illustrated in Figure~\ref{fig:half_column_analysis}, we visualize the spatial relationship among nodes, as shown in subfigure (a). 
To illustrate the spatial relationship, we calculate Dynamic Time Warping (DTW)~\cite{sakoe1971dynamic} similarity of traffic nodes in the PEMS08 dataset under different time intervals, \eg 5-min interval in subfigure (b) and 40-min interval in subfigure (c).

At a fine scale in Figure~\ref{fig:half_column_analysis} (b), spatial dependencies are dominated by physical proximity. For instance, adjacent Nodes 74 and 75 exhibit high similarity, while distant nodes show weak correlations.
Conversely, at a coarse scale in Figure~\ref{fig:half_column_analysis} (c), macroscopic functional similarities emerge. Geographically distant nodes (\eg Nodes from 80 to 115) that appear unrelated in the fine-scale view exhibit strong correlations. This is likely because they share similar functional roles (\eg suburban arterials) and synchronize during tidal commuting hours.
Existing methods, by ignoring this ``granularity-dependent'' nature of spatial topology, fail to capture these diverse interactions simultaneously. 
They either overfit local physical neighbors or miss the distant functional correlations that only emerge at coarser temporal scales.

Beyond spatial scaling, existing GCN-based spatial learners face a fundamental challenge in processing signal composition. 
Real-world traffic data is a superposition of globally shared patterns (\eg daily commuting fluctuation) and unique local variations (\eg accidents or sudden congestion). 
While shared patterns facilitate generalization, unique disturbances are crucial for precise node-level forecasting. However, standard GCNs propagate information through neighborhood aggregation, which tends to mix these heterogeneous signals into an entangled representation. 
Consequently, node representations often become biased toward dominant shared trends, causing unique, transient dynamics to be ``drowned out.'' This over-smoothing phenomenon degrades the model's ability to capture fine-grained heterogeneity, which is essential for robustness against non-stationary events.

To address these challenges, we propose a unified framework \name that synergizes multi-scale structural modeling with prototype-based residual decomposition.
First, to resolve the scale-effect issue, we introduce a Multi-Scale Masked Attention (MSMA) mechanism. By downsampling temporal features into multiple resolutions, MSMA generates granularity-specific spatial masks. 
This allows the model to explicitly attend to physical neighbors at fine scales and functional correlations at coarse scales, as observed in our preliminary analysis.
Second, to resolve the signal entanglement, we design a Prototype Residual Learning (PRL) module. 
Instead of simple aggregation, we maintain a momentum-updated prototype bank to encode global commonalities. 
We then adopt a ``subtract-to-refine'' strategy: extracting unique node representations by subtracting the prototype-aligned context from the node’s features. 
This explicitly preserves local deviations against over-smoothing.
To ease the reproducibility, we release our implementation code online\footnote{\url{https://github.com/Jlu-zcy/M2Net}}.

Our main contributions are summarized as follows:
\begin{itemize}
\item We identify that spatial correlations are not one-fold but vary significantly with temporal granularity. We validate that coarse-scale observations reveal long-range functional dependencies often missed by fine-scale views.
\item We propose \name, which integrates Multi-Scale Masked Attention to capture dynamic spatial structures and a Prototype Residual Learning module to explicitly disentangle unique local variations from global trends.
\item Extensive experiments on 5 real-world datasets demonstrate that \name achieves state-of-the-art performance, validating the effectiveness of coupling scale-specific spatial views with residual-based signal decomposition.

\end{itemize}

%% file: sections/4_Method.tex
\section{Methodology}

\begin{figure*}[!t]
    \centering
    \includegraphics[width=1.0\textwidth]{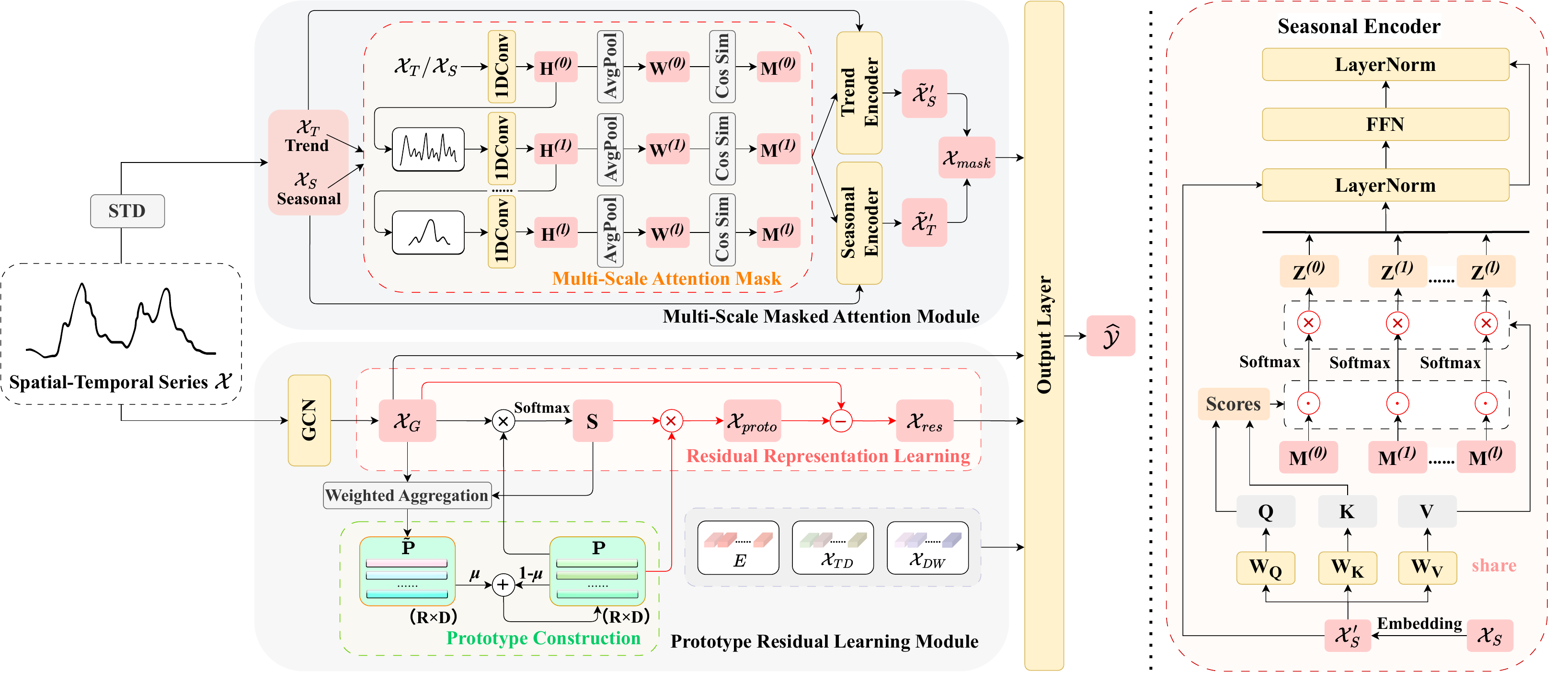}
    \caption{The framework of \name. Our model consists of two complementary branches: (1) Multi-Scale Masked Attention (MSMA) first applies Seasonal-Trend Decomposition (STD) to input time series, then captures spatial dependencies at multiple temporal resolutions by generating scale-specific masks from multi-resolution temporal features, and (2) Prototype Residual Learning (PRL) separates global patterns from node-specific variations using a momentum-updated prototype bank $\mathbf{P}$. Both branches are fused for final forecasting. The right panel shows the Seasonal Encoder (the trend encoder follows the same framework), where multi-scale masks $\{\mathbf{M}^{(l)}\}$ modulate shared QKV projections to produce scale-specific attention outputs $\{\mathbf{Z}^{(l)}\}$, aggregated through LayerNorm and FFN layers.
    } 
    \vspace{-3mm}
    \label{fig:framework}
\end{figure*}

\subsection{Problem Definition}
\noindent\textit{\textbf{Spatial-Temporal Forecasting.}}
We denote the road network as a graph $\mathcal{G} = (\mathcal{V}, \mathbf{A})$, where $\mathcal{V}$ denotes the set of nodes, and each node $v_i \in \mathcal{V}$ corresponds to a traffic sensor that continuously records traffic flow data. 
The network consists of $N = |\mathcal{V}|$ sensors, whose spatial dependencies are encoded by the adjacency matrix $\mathbf{A} \in \mathbb{R}^{N \times N}$.  
Each node $v_i \in \mathcal{V}$ is associated with a multivariate traffic observation sequence $\mathcal{X}_i = \{x_i^{t}\}_{t=1}^{T_{\text{total}}}$, where $x_i^{t} \in \mathbb{R}^{C}$ denotes the $C$-dimensional traffic state observed at time step $t$. 

Given the historical observations within a look-back window $[t-T+1,\, t]$, the goal of \textbf{spatial--temporal forecasting} is to predict the future traffic states of all nodes over the subsequent $T'$ time steps, formulated as:
\begin{equation}
    \widehat{\mathcal{Y}} = \mathcal{F}\big(\mathcal{X}_{t-T+1:t}, \mathbf{A}; \theta \big),
\end{equation}
where $\mathcal{F}(\cdot)$ denotes a learnable forecasting function parameterized by $\theta$, and the forecasting output $\widehat{\mathcal{Y}} \in \mathbb{R}^{N \times T' \times C'}$.

\subsection{Framework Overview}
As illustrated in Figure~\ref{fig:framework}, we propose a unified framework termed \name to jointly model scale-specific spatial structures and semantic variations in spatial-temporal forecasting.

The framework consists of two parallel and complementary branches. The Multi-Scale Masked Attention (MSMA)  (Section~\ref{subsec: MSMA}) directly takes the raw input sequences and captures spatial dependencies under multiple temporal scales by generating scale-specific attention masks. 
In parallel, the Prototype Residual Learning (PRL) (Section~\ref{subsec: PRL}) 
extracts global shared patterns via a prototype bank and models residual, node-specific variations by removing these common components. 
The outputs from the two branches are then fused to produce the final forecasting through the output layer (Section~\ref{subsec: Output}). 
The entire model is trained end-to-end, with the training objective and procedure defined in Section~\ref{subsec: Opt}, and the complexity analysis provided in Section~\ref{subsec:complex}.
 
\subsection{Multi-Scale Masked Attention Module}
\label{subsec: MSMA}
Spatial dependencies in spatial-temporal data vary across temporal resolutions: fine-grained scales mainly reflect local interactions, while coarse-grained scales capture long-range and functional correlations. Effectively modeling such heterogeneous spatial structures therefore requires explicitly accounting for temporal scale variations.
To this end, we propose a multi-scale dependency module to capture the evolution of spatial dependencies after temporal decomposition. 
Specifically, it first constructs spatial masks from corresponding multi-resolution temporal features (Section~\ref{subsubsec: Multi-Sacle}), and then integrates these heterogeneous dependency patterns through a mask-guided attention mechanism with shared parameters across scales (Section~\ref{subsubsec: SADM}).
This design enables unified and efficient modeling of spatial interactions across different temporal granularities.

\subsubsection{\textbf{Multi-Scale Attention Mask}}
\label{subsubsec: Multi-Sacle}
Temporal dynamics in real-world systems exhibit multi-scale characteristics, where dependencies manifest differently across various temporal granularities. Conventional attention mechanisms~\cite{vaswani2017attention}, however, apply uniform receptive fields across all time steps, failing to capture this scale-dependent nature of temporal interactions. 
To address this limitation, we devise MSMA in this section, which adaptively modulates the temporal attention scope according to different granularities, enabling the model to capture multi-resolution temporal interactions.

To facilitate effective temporal modeling, we first apply Seasonal-Trend Decomposition (STD), a proven strategy for modeling temporal dynamics~\cite{cao2025spatiotemporal, wu2021autoformer, qiu2025duet, zeng2023transformers}. 
Specifically, we first decompose the input time series $\mathcal{X}$ into seasonal $\mathcal{X}_S \in \mathbb{R}^{N \times T}$ and trend components $\mathcal{X}_T\in \mathbb{R}^{N \times T}$ using a moving average ~\cite{zeng2023transformers}. 
By separating time series into components with distinct behaviors, such decomposition facilitates more targeted modeling of long-term trends and short-term fluctuations.
These components provide complementary views for multi-scale modeling: $\mathcal{X}_S$ characterizes high-frequency variations, while $\mathcal{X}_T$ captures low-frequency trends.

Next, given the decomposed components $\mathcal{X}_S, \mathcal{X}_T$, we apply a sequence of strided one-dimensional convolutions (1DConv) \cite{kiranyaz20211d} along the temporal dimension to progressively downsample the temporal resolution and obtain the temporal representation at different granularities.
Without loss of generality, we describe the procedure using the seasonal component $\mathcal{X}_S$, and the same process is applied to $\mathcal{X}_T$. 

Starting from $\mathbf{H}^{(0)}=\mathcal{X}_S \in \mathbb{R}^{N \times T}$, we apply a sequence of $L$ strided 1DConv. 
At each scale $l$, the convolution layer with stride $s$ downsamples the temporal dimension while embedding the input into a feature space of dimension $F$:
\begin{equation}
\mathbf{H}^{(l)} = \text{1DConv}_{s}^{(l)}\left(\mathbf{H}^{(l-1)}\right) \in \mathbb{R}^{N \times F \times T^{(l)}},
\end{equation}
where $T^{(l)} = \lfloor T^{(l-1)}/s \rfloor$ and $l \in \{1, 2, \ldots, L\}$.
As $l$ increases, the temporal resolution of $\mathbf{H}^{(l)}$ is progressively reduced, leading to large effective temporal receptive fields. 
{For example, given an input sequence $\mathcal{X}_S \in \mathbb{R}^{N\times T}$ of length $T=12$ and stride $2$, the representations at scales $l \in \{0,1,2 \}$ correspond to temporal resolutions $T^{(l)} \in \{12, 6, 3 \}$, respectively, thereby capturing dependencies from fine-grained local patterns to coarse-grained global contexts.}

To derive scale-specific structural relationships between nodes, we construct a compact representation for each node at each temporal resolution by applying global average pooling along the temporal dimension:
\begin{equation}
    \mathbf{W}^{(l)} = \text{AvgPool}_T\left(\mathbf{H}^{(l)} \right) \in \mathbb{R}^{N \times F},
\end{equation}
which aggregates $\mathbf{H}^{(l)} \in \mathbb{R}^{N \times F \times T^{(l)}}$ into a fixed-length representation by averaging over the temporal dimension. Each $\mathbf{W}^{(l)}$ summarizes node-level temporal dynamics at scale $l$, enabling the measurement of structural similarity across nodes.

Based on these representations, we construct scale-specific attention masks that explicitly encode node interaction patterns under different temporal resolutions. 
For each scale $l$, we compute a pairwise similarity matrix between nodes using cosine similarity based on their scale-specific representations:
\begin{equation}
    \mathbf{\widehat{M}}^{(l)}_{ij} = \frac{\mathbf{W}^{(l)}_i \cdot \mathbf{W}^{(l)}_j}{\|\mathbf{W}^{(l)}_i\| \|\mathbf{W}^{(l)}_j\|}.
\end{equation}

The similarity scores are linearly mapped to $[0,1]$ as Bernoulli parameters 
$\mathbf{\overline{M}}^{(l)}_{ij} = (\mathbf{\widehat{M}}^{(l)}_{ij} + 1)/2$. 
Rather than directly using the continuous probability matrix, we perform Bernoulli resampling to obtain a binary attention mask $\textbf{M}^{(l)}$:
\begin{equation}
    \mathbf{M}^{(l)} \sim \mathrm{Bernoulli}\left( \mathbf{\overline{M}}^{(l)} \right), \quad \mathbf{M}^{(l)} \in \mathbb{R}^{N \times N}.
\end{equation}

A higher probability $\overline{\mathbf{M}}^{(l)}_{ij}$ implies a stronger interaction between node $i$ and node $j$ under scale $l$. 
To enable end-to-end optimization, we adopt the Gumbel–Softmax reparameterization trick~\cite{jang2017categorical} during the Bernoulli sampling process.

\subsubsection{\textbf{Structure-Aware Dependency Modeling}}
\label{subsubsec: SADM}
Given the scale-specific attention masks, we introduce a structure-aware dependency modeling mechanism to integrate heterogeneous interaction patterns into a unified spatial-temporal representation. 
For each decomposed component (seasonal or trend), we obtain a set of binary attention masks $\{\mathbf{M}^{(l)}\}_{l=0}^{L}$, where each mask encodes node interaction structures induced by a specific temporal scale.
Instead of instantiating independent attention layers for different scales, we design a mask-guided attention scheme that shares query–key–value (QKV) projections across scales, while modulating attention weights through scale-specific masks.
This design not only reduces parameter redundancy but also enforces consistent semantic alignment across temporal resolutions.

In the following, we use the seasonal component $\mathcal{X}_S$ to illustrate the structure-aware dependency modeling pipeline, and the trend component $\mathcal{X}_T$ follows the same procedure.
Given the decomposed seasonal component $\mathcal{X}_S \in \mathbb{R}^{N \times T}$, we first project it into a latent feature space to obtain embeddings $\mathcal{X}'_S = \mathcal{X}_S \mathbf{W}_S + \mathbf{b}_S$, 
where $\mathcal{X}'_S \in \mathbb{R}^{N \times D}$ is the embedded representation, $\mathbf{W}_S \in \mathbb{R}^{T \times D}$ and $\mathbf{b}_S \in \mathbb{R}^{D}$ are learnable parameters, and $D$ is the embedding dimension. 

To model node dependencies at different temporal resolutions, we apply a masked self-attention mechanism guided by the scale-specific masks $\{\mathbf{M}^{(l)}\}_{l=0}^{L}$. Specifically, we first compute the query, key, and value matrices as: 
$\mathbf{Q} = \mathcal{X'}_S\mathbf{W}_Q, \mathbf{K} = \mathcal{X'}_S\mathbf{W}_K, \mathbf{V} = \mathcal{X'}_S\mathbf{W}_V$, where $\mathbf{W}_Q, \mathbf{W}_K, \mathbf{W}_V \in \mathbb{R}^{D \times D}$ are learnable projection matrices. 
Notably, these QKV projections are computed once and shared across all temporal scales. 
Different scales are distinguished solely through their corresponding masks, which explicitly constrain the attention scope according to scale-induced interaction structures. 
The multi-scale structure-aware attention output is obtained by aggregating the masked attention responses across all scales:
\begin{equation}
    \mathbf{Z} = \sum_{l=0}^{L}\mathrm{Softmax}\!\left(\frac{\mathbf{Q}\mathbf{K}^\top\odot\mathbf{M}^{(l)}
    +(1-\mathbf{M}^{(l)})\cdot (-C)}{\sqrt{d}}\right)\mathbf{V},
\end{equation}
{where $\odot$ denotes element-wise multiplication and $C\gg 0$ is a sufficiently large constant used to suppress masked positions, resulting in near-zero attention weights after the softmax operation.}
This design captures diverse spatial dependency patterns across temporal scales while maintaining computational efficiency by sharing the QKV projections across scales.

Following the standard transformer~\cite{vaswani2017attention}, the aggregated features are passed through a transformer-style encoder block with residual connections and layer normalization~\cite{ba2016layer} for stable optimization and effective feature propagation. 
A position-wise feed-forward network~\cite{bebis2002feed} is then applied to enhance representation capacity. The final seasonal representation $\tilde{\mathcal{X'}}_{S} \in \mathbb{R}^{N \times D}$ is obtained as: 
\begin{equation}
\begin{aligned}
\mathbf{Z}' &= \mathrm{LayerNorm}(\mathcal{X'}_S + \mathbf{Z}), \\
\tilde{\mathcal{X'}}_{S} &= \mathrm{LayerNorm}\Big(\mathbf{Z}' + \mathrm{FFN}(\mathbf{Z}')\Big).
\end{aligned}
\end{equation}

We apply the above structure-aware attention mechanism independently to both the seasonal component $\mathcal{X}_S$ and trend component $\mathcal{X}_T$ using separate transformer-style encoders, yielding $\tilde{\mathcal{X'}}_{S}$ and $\tilde{\mathcal{X'}}_{T} \in \mathbb{R}^{N \times D}$. This allows each component to focus on its own characteristic dependency patterns.
The outputs of the two branches are finally combined to form the overall mask-aware representation:
\begin{equation}
    \mathcal{X}_{mask} = \tilde{\mathcal{X'}}_{S} + \tilde{\mathcal{X'}}_{T}.
\end{equation}

\subsection{Prototype Residual Learning Module}
\label{subsec: PRL}
In real-world spatial-temporal systems, node representations often entangle globally shared patterns, \eg common traffic fluctuations or seasonal trends, with node-specific variations, \eg local anomalies or node-specific deviations. 
Such entanglement obscures fine-grained local dynamics and limits the model’s ability to adapt to node heterogeneity.
To address this, we propose a Prototype Residual Learning (PRL) module, which explicitly separates common spatial-temporal patterns from node-unique deviations. 
PRL maintains a momentum-updated prototype bank to represent global patterns (Section~\ref{subsubsec: PC}) and removes them from node representations to obtain residual features that emphasize node-specific variations (Section~\ref{subsubsec: RRL}), encouraging the model to focus on informative local dynamics while retaining global consistency.

\subsubsection{\textbf{Prototype Construction}}
\label{subsubsec: PC}
Before constructing prototypes, we derive node-level latent representations from historical time series $\mathcal{X}$. 
To enrich temporal semantics, we augment the input with periodic temporal context, including time-of-day and day-of-week encodings~\cite{shao2022spatial}.
Each temporal attribute is normalized by its period and mapped to $[0,1]$, 
preserving periodic structure while ensuring a consistent numerical scale. This augmentation provides fine-grained temporal cues without introducing scale imbalance.


The augmented input is reshaped into $\mathcal{X} \in \mathbb{R}^{N \times (T \cdot (C+2))}$ and projected into a latent space using a graph convolutional embedding module. 
To aggregate spatial context, we adopt an adaptive graph convolution based on Chebyshev polynomials~\cite{mason2002chebyshev} that captures multi-hop spatial dependencies.
{Concretely, node-level embeddings are computed as:
\begin{equation}
\mathcal{X}_G=
\sum_{k=0}^{K-1}
T_k(\mathbf{A})\mathcal{X}\mathbf{W}_k,
\end{equation}
where $T_k(\cdot)$ denotes the $k$-th order Chebyshev polynomial and $\mathbf{W}_k$ represents the learnable weight matrix associated with the $k$-th Chebyshev order. The adaptive adjacency matrix is given by $\mathbf{A}=\text{Softmax}\left(\text{ReLU}(\mathbf{E}\mathbf{E}^{\top})\right)\in\mathbb{R}^{N\times N},$ where $\mathbf{E}\in\mathbb{R}^{N\times D}$ denotes learnable node embeddings.}

The resulting node-level embeddings $\mathcal{X}_{G} \in \mathbb{R}^{N \times D}$ serve as the basis for prototype construction.
We maintain a prototype bank $\mathbf{P} = \{\mathbf{p}^1, \dots, \mathbf{p}^R\}$ with $R$ prototypes, where each prototype $\mathbf{p}^r \in \mathbb{R}^{D}$ represents a globally shared spatial-temporal pattern across nodes. 
Node-level embeddings $\mathcal{X}_{G}$ are softly associated with prototypes by computing the scaled inner-product similarity: $\mathbf{S} = \mathrm{Softmax}\!\left(\frac{\mathcal{X}_{G} \mathbf{P}^{\top}}{\tau \sqrt{D}}\right)$, 
where $\mathbf{S} \in \mathbb{R}^{N \times R}$ denotes the soft assignment weights and $\tau$ is the temperature parameter controlling assignment sharpness.

Using these assignment weights, each prototype is updated as a weighted aggregation of node embeddings: 
\begin{equation}
    \tilde{\mathbf{p}}^r = \frac{\sum_{n=1}^{N} \mathbf{S}^{n,r}\, \mathcal{X}_{G}^{n}}{\sum_{n=1}^{N} \mathbf{S}^{n,r}}.
\end{equation}

Rather than updating the prototypes via back-propagation, we adopt an exponential moving average (EMA) strategy~\cite{tarvainen2017mean}:
\begin{equation}
    \mathbf{P} = \mu \mathbf{P} + (1 - \mu)\tilde{\mathbf{P}},
\end{equation}
where $\mu$ is momentum coefficient controlling the update rate.
This EMA-based approach decouples prototype updates from batch-wise gradient noise, 
enabling the prototype bank to gradually capture stable and representative global patterns across training iterations rather than overfitting to batch fluctuations.

\subsubsection{\textbf{Residual Representation Learning}}
\label{subsubsec: RRL}
With the updated prototypes, we retrieve prototype representations for each node to disentangle shared patterns from unique node-specific variations. 
Given soft assignment weights $\mathbf{S}$ and prototypes $\mathbf{P}$, the
prototype representation $\mathcal{X}_{proto}$ for node $n$ is computed as: 
\begin{equation}
    \mathcal{X}_{proto}^n = \sum_{r=1}^{R} \mathbf{S}^{n,r} \, \mathbf{p}^r, \quad n = 1, \dots, N.
\end{equation}

The residual representation $\mathcal{X}_{res}$ for each node is then obtained by subtracting the prototype representation $\mathcal{X}_{proto}$ from the original node-level embedding $\mathcal{X}_{G}$:
\begin{equation}
    \mathcal{X}_{res}^n = \mathcal{X}_{G}^{n} - \mathcal{X}_{proto}^n, \quad n = 1, \dots, N.
\end{equation}

This residual operation explicitly removes globally shared spatial–temporal trends, forcing the model to focus on node-specific deviations that are critical for capturing fine-grained local dynamics. 
The resulting residual representations are subsequently fed into downstream modules for forecasting.


\subsection{Output Layer}
\label{subsec: Output}

The output layer transforms the fused hidden representations into multi-step forecasts for each node. 
We integrate heterogeneous information, which preserves spatial and temporal identities, and yield a comprehensive representation for forecasting.
Specifically, we combine representations from the graph-based node-level embeddings $\mathcal{X}_G$, prototype residual features $\mathcal{X}_{res}$, and multi-scale masked attention module output $\mathcal{X}_{mask}$, together with lightweight identity embeddings that encode node, \ie $E$ and temporal context, \ie $\mathcal{X}_{TD}$ and $\mathcal{X}_{DW}$, where the temporal embeddings are indexed by the timestamp of the last step in the input time series to represent the corresponding time-of-day and day-of-week.
These components jointly form a comprehensive hidden representation $\mathcal{X}_{hid}$:


\begin{equation}
    \mathcal{X}_{hid} = \mathcal{X}_{G} || \mathcal{X}_{res} || \mathcal{X}_{mask} || E || \mathcal{X}_{TD} || \mathcal{X}_{DW}.
\end{equation}

The fused representation $\mathcal{X}_{hid} \in \mathbb{R}^{N \times 6D}$ is then processed by stacked MLP blocks with residual connections to integrate information from different sources, followed by a 1DConv regression head to produce the final forecasting outputs:
\begin{equation}
    \widehat{\mathcal{Y}} = \text{1DConv}(\text{MLP}(\mathcal{X}_{hid})+\mathcal{X}_{hid}).
\end{equation}

\begin{algorithm}[!t]
\caption{The $M^2Net$ Framework}
\label{alg}
\raggedright
\begin{algorithmic}[1]
\State \textbf{Input:} Historical observations $\mathcal{X}$
\State \textbf{Output:} Predicted future values $\widehat{\mathcal{Y}}$  

\State Initialize prototype bank $\mathbf{P}$

\While{not converge}

\State Sample a mini-batch $\mathcal{X}$

\State $\mathcal{X}_{mask}\leftarrow \mathrm{MSMA}(\mathcal{X})$
\Comment{Eq.~(2)--(8)}

\State $\mathcal{X}_G\leftarrow \mathrm{GCN}(\mathcal{X})$
\Comment{Eq.~(9)}

\State Update prototype bank $\mathbf{P}$ via EMA
\Comment{Eq.~(10--11)}

\State $\mathcal{X}_{res}\leftarrow \mathrm{PRL}(\mathcal{X}_G,\mathbf{P})$
\Comment{Eq.(12)--(13)}

\State $\mathcal{X}_{hid}\leftarrow
\mathrm{Concat}(\mathcal{X}_G,
\mathcal{X}_{res},\mathcal{X}_{mask},E,
\mathcal{X}_{TD},\mathcal{X}_{DW})$
\Comment{Eq.~(14)}

\State $\widehat{\mathcal{Y}}
\leftarrow \mathrm{Predictor}(\mathcal{X}_{hid})$
\Comment{Eq.~(15)}

\State Compute loss $\mathcal{L}$
\Comment{Eq.~(16)}

\State Update model parameters $\theta$ via back-propagation

\EndWhile

\State \Return $\widehat{\mathcal{Y}}$

\end{algorithmic}
\end{algorithm}
\vspace{-2mm}

\subsection{Optimization Objectives}
\label{subsec: Opt}
We minimize the Mean Absolute Error (MAE) between the predicted future values $\widehat{\mathcal{Y}}$ and the ground-truth future observations $\mathcal{Y}$ as the training loss:
\begin{equation}
\mathcal{L} = \frac{1}{|\Omega|} \sum_{(t,n)\in\Omega} \left| \mathcal{\widehat{Y}}_{t,n} - \mathcal{Y}_{t,n} \right|,
\end{equation}
where $\Omega$ represents the set of all node-time pairs in the forecasting horizon.

The forecasting network is optimized using the MAE objective, while the prototype bank is updated via an EMA mechanism based on node-level embeddings. This update strategy allows the prototype bank to progressively capture stable global patterns across training iterations without gradients through backpropagation, thereby ensuring training stability.

{The overall training procedure of the $M^2Net$ is summarized in Algorithm~\ref{alg}. Given the historical observations, the model first extracts multi-scale spatial dependencies through the MSMA module (line 6). Meanwhile, graph-based node-level representations are obtained by the GCN module (line 7). The PRL module then captures globally shared patterns via the updated prototype bank (line 8) and derives node-specific residual representations by removing prototype components from node-level embeddings (line 9). Subsequently, the outputs of MSMA and PRL are concatenated with node-level embeddings and temporal identity embeddings to form the hidden representation for forecasting (lines 10--11). Finally, the model parameters are optimized using the forecasting loss, while the prototype bank is updated through an exponential moving average strategy until convergence (lines 12--13).}

\subsection{Complexity Analysis}
\label{subsec:complex}
The computational cost of \name mainly comes from three components:
\textbf{\textit{(1)}} \textit{\textbf{Multi-Scale Mask Construction.}}
For each scale, temporal features are extracted by strided 1D convolutions and then aggregated through average pooling. The resulting node representations are used to compute pairwise cosine similarities for mask generation. The complexity is $O(NFT)+O(LN^2F)$. 
\textbf{\textit{(2)}} \textit{\textbf{Multi-Scale Masked Attention (MSMA).}}
Given the scale-specific masks, attention is computed using shared QKV projections. For each scale, the attention operation requires $O(N^2D)$. Aggregating across L scales yields $O(LN^2D)$. 
\textbf{\textit{(3)}} \textit{\textbf{Prototype Residual Learning (PRL).}} 
The adaptive graph convolution used for node embedding incurs $O(N^2D)$, while prototype assignment and prototype aggregation require $O(NRD)$. Since $R\ll N$, the prototype-related overhead is negligible compared with graph propagation and attention computation. Therefore, the overall complexity of \name can be expressed as $O(NFT+LN^2F+LN^2D+N^2D+NRD)$. Since D is typically larger than $F$ and $R\ll N$, the dominant complexity is $O((L+1)N^2D)$. Although the theoretical complexity is quadratic with respect to the number of nodes, the number of scales $L$ is small (typically $L=3$), and the prototype bank size $R$ is much smaller than $N$. Therefore, \name remains computationally efficient in practice, which is further validated by the runtime results reported in Section~\ref{sub:computation} and Table~\ref{tab:computational}.

%% file: sections/5_Experiments.tex
\begin{table}[!t]
\centering
\caption{Dataset statistics.}
\begin{tabular}{cccc}
\toprule
\textbf{Dataset} & \textbf{Sensors} & \textbf{Timesteps} & \textbf{Time Range} \\
\midrule
PEMS-BAY & 325 & 52,116 & 01/2017 -- 06/2017 \\
PEMS07 & 883 & 28,224 & 05/2017 -- 08/2017 \\
PEMS08 & 170 & 17,856 & 07/2016 -- 08/2016 \\
PEMS03 & 358 & 26,208 & 09/2018 -- 11/2018 \\
PEMS04 & 307 & 16,992 & 01/2018 -- 02/2018 \\
\bottomrule
\end{tabular}
\vspace{-6mm}
\label{tab:dataset}
\end{table}

\section{Experiments}
In this section, we first introduce the experimental setup (Section~\ref{sub:setups}) and report the overall comparison results (Section~\ref{sub:overall}). Ablation studies are presented in Sections~\ref{sub:Ablation}, followed by analyses of the compatibility and effectiveness of the PRL module in Section~\ref{sub:PRL_compatibilty} and~\ref{sub:PRL_effect}. Finally, we investigate hyper-parameter sensitivity (Section~\ref{sub:hyper}), computational efficiency (Section~\ref{sub:computation}), MSMA generalization analysis (Section~\ref{sub:MSMA gene}), multi-scale mask complementarity analysis (Section~\ref{sub:Mask_com}) and scalability evaluation on larger graphs (Section~\ref{sub:Large}).

\subsection{Experimental Setups}
\label{sub:setups}

\subsubsection{Datasets}
\label{subsub:Datasets}
We evaluate our model on five widely used traffic datasets: PEMS-BAY~\cite{li2018dcrnn_traffic} and PEMS03/04/07/08~\cite{song2020spatial}. PEMS03/04/07/08 contain traffic flow measurements collected from the California Performance Measurement System, while PEMS-BAY records traffic speed data from loop detectors in the Los Angeles and Bay Area regions. All datasets are sampled at 5-minute intervals. Detailed statistics are summarized in Table~\ref{tab:dataset}. Following prior studies~\cite{shao2024exploring}, PEMS-BAY is split into 70\%/10\%/20\% for training/validation/testing, while PEMS03/04/07/08 use a 60\%/20\%/20\% split.

\begin{table*}[!t]
\centering
\small
\caption{Performance comparison of different methods. ``(ours)'' means our proposed method \name. The best results are \textbf{bolded} and the second-best results are \underline{underlined}. 
``(a)'' denotes Transformer-based methods, 
``(b)'' denotes GRU-based methods, 
``(c)'' denotes CNN/GNN-based baselines, 
``(d)'' denotes lightweight methods, 
``(ours)'' denotes \name.
}
\renewcommand{\arraystretch}{1.0}
\resizebox{\textwidth}{!}{%
\begin{tabular}{c c ccc ccc ccc ccc ccc}
\toprule
&\textbf{Method}
& \multicolumn{3}{c}{\textbf{PEMS-BAY}}
& \multicolumn{3}{c}{\textbf{PEMS07}}
& \multicolumn{3}{c}{\textbf{PEMS08}}
& \multicolumn{3}{c}{\textbf{PEMS03}}
& \multicolumn{3}{c}{\textbf{PEMS04}}

\\
\midrule
&\textbf{Metrics}
& MAE & RMSE & MAPE(\%)
& MAE & RMSE & MAPE(\%)
& MAE & RMSE & MAPE(\%)
& MAE & RMSE & MAPE(\%)
& MAE & RMSE & MAPE(\%) \\
\midrule
\multirow[c]{3}{*}{(a)}
& PDFormer
& 1.62 & 3.65 & 3.66
& 19.97 & 32.95 & 8.55
& \underline{14.09} & 23.41 & \textbf{9.05}
& 14.94 & 25.39 & 15.82
& \underline{18.36} & 30.03 & \textbf{12.00} 
\\

& STAEformer
& 1.65 & 3.61 & 3.75
& 20.09 & 33.36 & 8.41
& 14.17 & 23.63 & 9.28
& 15.27 & 26.76 & 15.88
& 18.34 & 31.14 & 13.06
\\

& STDN
& 1.60 & 3.78 & 3.62
& 22.94 & 36.06 & 10.32
& 14.41 & 23.89 & 10.26
& 15.34 & 27.12 & 17.71
& 18.79 & 34.13 & 16.96
\\

\midrule
\multirow{3}{*}{(b)}

& DGCRN
& 1.64 & 3.67 & 3.66
& 20.24 & 33.08 & 8.71
& 14.34 & 23.53 & 9.48
& 15.09 & 26.02 & 16.05
& 18.68 & 30.21 & 13.04 
\\

& AGCRN
& 1.65 & 3.68 & 3.75
& 20.40 & 34.24 & 8.62
& 15.54 & 24.77 & 10.15
& 15.60 & 26.88 & 15.25
& 19.25 & 31.10 & 13.00
\\

& $D^2$STGNN
& 1.62 & 3.69 & 3.68
& 20.33 & 33.23 & 8.73 
& 14.36 & 23.46 & 9.32
& 14.84 & 25.41 & \textbf{15.17}
& 18.61 & 30.13 & 12.82
\\

\midrule
\multirow{5}{*}{(c)}
& STGCN
& 1.74 & 3.76 & 4.06
& 21.89 & 35.64 & 9.45
& 15.72 & 24.93 & 10.64
& 17.04 & 29.62 & 17.36 
& 19.27 & 30.83 & 13.16
\\

& MTGNN
& 1.63 & 3.66 & 3.62
& 20.97 & 34.20 & 8.90
& 15.22 & 24.09 & 9.89
& 15.31 & 25.95 & 15.24
& 19.20 & 31.81 & 13.26
\\

& GWNet
& 1.61 & 3.61 & 3.63
& 19.92 & 35.84 & 8.62
& 14.59 & 23.57 & 9.41
& \textbf{14.55} & 25.66 & 15.50
& 18.80 & 30.33 & 12.92
\\

& STNorm
& 1.65 & 3.66 & 3.75
& 21.23 & 34.54 & 8.96
& 15.94 & 25.05 & 10.01
& 15.82 & 26.48 & \underline{15.18}
& 19.44 & 31.24 & 13.42
\\

& STWave
& 1.65 & 3.70 & 3.74
& 20.74 & 34.05 & 8.77
& 15.41 & 24.62 & 9.94
& 15.28 & 26.62 & 15.27
& 19.06 & 31.02 & 12.52
\\

\midrule
\multirow{4}{*}{(d)}
& STID
& 1.70 & 3.86 & 3.91
& 20.50 & 34.66 & 8.75
& 14.22 & 23.33 & 9.31
& 15.41 & 27.42 & 16.50
& 18.47 & 30.02 & 12.69
\\

& BigST
& 1.65 & \underline{3.58} & 3.77
& 20.31 & 33.57 & 8.57
& 14.19 & \underline{23.26} & 9.29
& 15.30 & 25.77 & 16.54 
& 18.42 & \underline{29.96} & 12.92
\\

& LightST
& 1.78 & 3.88 & 4.15
& 20.78 & 33.95 & 8.98 
& 15.43 & 24.52 & 9.84 
& 15.11 & \underline{24.74} & 15.41
& 19.21 & 31.31 & 12.70
\\

& RAST
& \underline{1.57} & 3.61 & \underline{3.55}
& \underline{19.52} & \underline{32.73} & \underline{8.32}
& 15.14 & 24.16 & 9.78
& 15.94 & 27.17 & 16.47
& 19.48 & 31.08 & 13.34

\\
\midrule
\multirow{1}{*}{(ours)}
& \textbf{\name}
& \textbf{1.52} & \textbf{3.49} & \textbf{3.41}
& \textbf{19.39} & \textbf{32.52} & \textbf{8.29}
& \textbf{14.02} & \textbf{23.10} & \underline{9.15}
& \underline{14.65} & \textbf{24.68} & \textbf{15.17}
& \textbf{18.20} & \textbf{29.89} & \underline{12.43}

 \\

\bottomrule
\end{tabular}
}
\vspace{-3mm}
\label{tab:baseline_comparison_grouped}
\end{table*}

\subsubsection{Baselines}
\label{subsub:Baselines}

To comprehensively evaluate \name, we compare it with sixteen representative state-of-the-art baselines using Mean Absolute Error (MAE), Root Mean Squared Error (RMSE), and Mean Absolute Percentage Error (MAPE). The baselines cover four categories: Transformer-based, GRU-based, CNN/GNN-based, and lightweight methods. 

\textit{\textbf{(a) Transformer-based methods:}}
\textbf{PDFormer}~\cite{jiang2023pdformer}: Models dynamic spatial dependencies and propagation delays through spatial attention and delay-aware feature transformation.
\textbf{STAEformer}~\cite{liu2023spatio}: Uses adaptive spatial-temporal embeddings with Transformers to capture spatial-temporal and chronological patterns.
\textbf{STDN}~\cite{cao2025spatiotemporal}: Learns dynamic graphs and separately models trend and seasonal components.

\textit{\textbf{(b) GRU-based methods:}}
\textbf{DGCRN}~\cite{li2023dynamic}: Generates dynamic graph filters with hyper-networks and integrates them with static graphs.
\textbf{AGCRN}~\cite{bai2020adaptive}: Learns node-specific patterns and graph structures without predefined graphs.
\textbf{$D^2$STGNN}~\cite{shao2022decoupled}: Separately models diffusion and inherent traffic signals with dynamic graph learning.

\textit{\textbf{(c) CNN/GNN-based methods:}}
\textbf{STGCN}~\cite{yu2018spatio}: Uses graph and temporal convolutions to capture spatial-temporal dependencies.
\textbf{MTGNN}~\cite{wu2020connecting}: Learns directed spatial relations and captures spatial-temporal dependencies with mix-hop propagation and dilated convolutions.
\textbf{GWNet}~\cite{wu2019graph}: Learns hidden spatial dependencies with adaptive graphs and models long sequences using dilated convolutions.
\textbf{STNorm}~\cite{deng2021st}: Uses spatial and temporal normalization to refine traffic representations.
\textbf{STWave}~\cite{fang2023spatio}: Decomposes traffic signals into stable trends and fluctuating events using a dual-channel architecture.

\textit{\textbf{(d) Lightweight methods:}}
\textbf{STID}~\cite{shao2022spatial}: Incorporates spatial and temporal identity information into a MLP-based predictor.
\textbf{BigST}~\cite{han2024bigst}: Uses scalable long-sequence modeling and linearized spatial convolution for large-scale traffic networks.
\textbf{LightST}~\cite{zhang2025efficient}: Distills spatial-temporal knowledge from high-capacity teachers into lightweight MLP students.
\textbf{RAST}~\cite{ruan2025rast}: Uses retrieval-augmented mechanisms and decoupled encoders to enhance spatial-temporal prediction.

\subsubsection{Implementation Details}
\label{subsub:Implementation Details}
Our framework is implemented in Python 3.12.3 and PyTorch 2.7.0. All experiments are conducted on a server equipped with NVIDIA RTX 3090 GPUs. The look-back window $T$ and the forecasting horizon $T'$ are set to 12. The model is trained using the Adam optimizer~\cite{Kingma2014AdamAM} with a batch size of 64. 
For all baselines, we use the official implementations and tune the key hyper-parameter to ensure optimal performance. We conduct repetitive runs as experimental results and report the average performance across all 12 forecasting steps as the final experimental results.

\subsection{Overall Performance}
\label{sub:overall}
The comparative results, summarized in Table~\ref{tab:baseline_comparison_grouped}, reveal several key insights:
\textbf{\textit{(1)}} Compared with transformer-based, GRU-based, and CNN/GNN-based methods, \name\ achieves consistently superior performance across all datasets. 
For example, on PEMS03, \name reduces MAE by 1.28\% compared to $D^2$STGNN, and achieves the best RMSE. 
On PEMS-BAY, our model reduces MAE by 5.0\% compared to STDN and achieves the lowest RMSE of 3.49. On PEMS07, it improves MAE by 2.9\% over PDFormer and reduces RMSE by 1.31\%. 
These improvements highlight the advantage of modeling spatial dependencies at multiple temporal resolutions. Existing approaches typically assume a single unified temporal scale, while our multi-scale masked attention module captures temporal scale-dependent spatial interactions. At fine temporal scales, the learned masks emphasize local physical correlations, whereas at coarser scales they highlight long-range functional relations.
\textbf{\textit{(2)}} The performance gains over lightweight models highlight the importance of disentangling globally shared patterns with node-specific variations. 
By removing shared global trends, our mechanism encourages the model to focus on residual representations that are more indicative of forecasting. Compared with BigST, our method reduces MAE by 4.25\% on PEMS03 and 1.19\% on PEMS04. 
These consistent improvements suggest that removing shared global trends enables the model to focus on informative residual representations and better capture local variations.
\textbf{\textit{(3)}} The performance improvements are particularly pronounced on datasets with heterogeneous spatial structures. For instance, PEMS-BAY (325 nodes) shows the smallest MAE reduction of 3.18\%,  
suggesting that our multi-scale modeling provides greater advantages in complex road networks where traffic patterns exhibit diverse spatial dependencies at different temporal granularities.
\textbf{\textit{(4)}} A particular observation is that \name achieves the best RMSE across all five datasets without exception. 
Since RMSE penalizes large forecasting errors through its squared term, this consistent advantage indicates that our model effectively mitigates extreme forecasting failures, which is crucial for safety-critical traffic applications.
\textbf{\textit{(5)}} While \name achieves the best MAE and RMSE on PEMS04, it ranks second in MAPE. 
This discrepancy stems from MAPE's sensitivity to low traffic volumes, where small absolute errors yield large percentage errors, suggesting our model excels at reducing absolute forecasting errors and indicating a metric-specific trade-off between absolute and relative accuracy.

\subsection{Ablation Study}
\label{sub:Ablation}
{To validate the contribution of each component in our proposed framework, we conduct comprehensive ablation studies on PEMS03 (358 nodes, 3 months), PEMS04 (307 nodes, 2 months), and PEMS08 (170 nodes, 2 months), which provide diverse combinations of network scale and temporal coverage, ensuring robust validation across different traffic scenarios. The results are summarized in Table~\ref{tab:ablation}.}

\textbf{\textit{(1)}} \textbf{Effectiveness of multi-scale masked attention module.} 
\ding{172} We first examine the contribution of the multi-scale masking mechanism by comparing several variants. The variant \textbf{w/o MSMA} removes the entire multi-scale masked attention module, resulting in consistent performance degradation across all datasets. This demonstrates the importance of modeling spatial dependencies at multiple temporal scales, as different temporal granularities reveal distinct spatial interaction patterns in traffic networks.
\ding{173} We further evaluate different masking strategies: \textbf{Mask-T} (masking only the trend component), \textbf{Mask-S} (masking only the seasonal component), and \textbf{w/o Dec}  (masking without prior decomposition). The results show that our full model with dual-branch masking consistently outperforms all variants, confirming the effectiveness of our multi-scale masking design. {Notably, \textbf{Mask-T} performs better on PEMS03 and PEMS08, while \textbf{Mask-S} shows advantages on PEMS04, suggesting that PEMS03 and PEMS08 have stronger trend-dominated patterns where long-term dependencies are critical, while PEMS04 exhibits more complex seasonal variations. These observations highlight that our multi-scale design effectively adapts to diverse dataset characteristics by integrating both fine-grained seasonal fluctuations and coarse-grained trend patterns.}
{\ding{174} We also compare our masking-based approach with standard attention mechanisms. The \textbf{A-Dual} variant applies standard Transformer attention to decomposed trend and seasonal components, while \textbf{A-Raw} applies standard attention without decomposition. Notably, \textbf{A-Raw} outperforms \textbf{A-Dual} on most datasets,  
suggesting that standard attention may not fully leverage decomposition without proper guidance. The decomposition may even add unnecessary complexity when combined with standard attention. In contrast, our masked attention explicitly learns which temporal granularities are important for capturing spatial dependencies, using decoupling mechanism and leading to efficient representations.}

\textbf{\textit{(2)}} \textbf{Impact of multi-scale masking.}
We evaluate the effect of the number of scales by comparing single-scale (\ding{172} \textbf{1-Scale}, \ie $\mathbf{M}^{(0)}$), two-scale (\ding{173} \textbf{2-Scale}, \ie $\mathbf{M}^{(0)},\mathbf{M}^{(1)}$), and three-scale (ours), \ie $\mathbf{M}^{(0)},\mathbf{M}^{(1)},\mathbf{M}^{(2)}$. Although adding an intermediate scale does not consistently improve every metric, the full three-scale configuration achieves the best overall performance. This confirms that hierarchical multi-scale modeling is beneficial for capturing spatial correlation under both fine-grained and coarse-grained temporal dynamics.

{\textbf{\textit{(3)}} \textbf{Effectiveness of stride selection.}
We further investigate the impact of the stride parameter $s$ used for temporal downsampling. The default setting $s=2$ constructs a three-level hierarchy ($12 \rightarrow 6 \rightarrow 3$), enabling the model to capture dependencies from fine-grained local patterns to coarse-grained global contexts. We additionally evaluate \textbf{$s=3$} and \textbf{$s=4$}. As shown in Table~\ref{tab:ablation}, $s=2$ consistently achieves the best overall performance across datasets, indicating that the hierarchy generated by $s=2$ provides a more effective balance between temporal resolution and receptive field expansion.}

\begin{table}[!t]
\centering
\setlength{\tabcolsep}{1pt}
\caption{Ablation study results.
}
\renewcommand{\arraystretch}{1.2}
\resizebox{\linewidth}{!}{%
\begin{tabular}{c ccc ccc ccc}
\toprule
\textbf{Method}
& \multicolumn{3}{c}{\textbf{PEMS03}}
& \multicolumn{3}{c}{\textbf{PEMS04}}
& \multicolumn{3}{c}{\textbf{PEMS08}}\\
\midrule
\textbf{Metrics}
& MAE & RMSE & MAPE(\%)
& MAE & RMSE & MAPE(\%)
& MAE & RMSE & MAPE(\%) \\
\midrule

\textbf{w/o MSMA}
& 14.96	& 25.65	& 15.70
& 18.33	& 30.25	& 12.56
& 14.13	& 23.27	& 9.24
\\

\textbf{Mask-T}
& \textbf{14.65} & 25.05 & 15.32
& 18.33	& 30.43	& 12.54
& 14.06	& 23.20	& 9.20
\\

\textbf{Mask-S}
& 14.77	& 25.72	& 16.32
& \underline{18.24}	& 30.05	& \textbf{12.39}
& 14.16	& 23.32	& 9.25
\\

\textbf{w/o Dec}
& 14.82	& 25.96	& 15.57
& 18.31	& 30.16	& 12.67
& 14.07	& 23.16	& 9.20
\\

\textbf{A-Dual}
& 14.90	& 25.64	& 15.33
& 18.30	& 30.17	& 12.50
& 14.11	& 23.29	& 9.23
\\

\textbf{A-Raw}
& 14.73	& 25.97	& 15.55
& 18.26	& \underline{30.03}	& 12.56
& 14.08	& 23.18	& 9.30
\\

\midrule

\textbf{1-Scale}
& 14.72	& \underline{25.01}	& 16.11
& 18.33	& 30.36	& 12.58
& 14.08	& 23.34	& 9.53
\\

\textbf{2-Scale}
& 14.71	& 25.75	& 15.69
& 18.30	& 30.24	& 12.49
& 14.09	& 23.19	& 9.44
\\

\midrule
\textbf{$s=3$}
& 14.75 & 25.52 & 15.84
& 18.29 & 30.23 & 12.58
& 14.08 & 23.21 & 9.36
\\

\textbf{$s=4$}
& 14.85 & 26.12 & 15.76
& 18.28 & 30.14 & 12.48
& \underline{14.05} & 23.18 & 9.29
\\

\midrule
\textbf{C-$\mathcal{X}_{proto}$}
& 14.70	& 26.82	& 15.62
& 18.31	& 30.16	& 12.53
& 14.13	& 23.26	& 9.35
\\

\textbf{w/o PRL}
& \underline{14.67}	& 25.36	& 15.47
& 18.36	& 30.33	& 12.48
& 14.06	& \underline{23.12}	& 9.31
\\

\midrule

\textbf{P $\rightarrow$ M}
& 14.72 & 26.59 & 15.36 
& 18.33 & 30.29 & 12.51
& 14.13 & 23.23 & 9.44
\\

\textbf{M $\rightarrow$ P}
& 14.71 & 25.96 & \underline{15.28} 
& 18.36 & 30.35 & 12.60
& \underline{14.05} & 23.14 & 9.45
\\

\midrule

\textbf{PerScale}
& 14.76 & 26.81 & 15.62 
& 18.30 & 30.29 & 12.50
& 14.12 & 23.29 & \underline{9.16}

\\

\textbf{ScaleRepr}
& 14.71 & 25.78 & 15.73 
& 18.33 & 30.29 & 12.48
& \underline{14.05} & 23.13 & 9.41
\\
\midrule

\textbf{\name}
& \textbf{14.65} & \textbf{24.68} & \textbf{15.17}
& \textbf{18.20} & \textbf{29.89} & \underline{12.43}
& \textbf{14.02} & \textbf{23.10} & \textbf{9.15}
\\

\bottomrule
\end{tabular}
}
\vspace{-6mm}
\label{tab:ablation}
\end{table}

\textbf{\textit{(4)}} \textbf{Effectiveness of prototype residual learning module.}
To validate the design of our prototype-based approach, we compare two variants: 
\ding{172} \textbf{C-$\mathcal{X}_{proto}$}, which directly concatenates prototype representations instead of residual features. It shows degraded performance, particularly evident in the RMSE metric on PEMS03 (26.82 vs. 24.68), suggesting that directly using prototypes introduces redundancy and fails to capture node-specific deviations from global patterns. 
\ding{173} \textbf{w/o PRL}, which completely removes the prototype residual module and also exhibits a performance decline. This confirms that incorporating node-aware information through prototype residuals is essential. 
These results validate our design choice of using residual connections to model the deviation between individual nodes and prototypes, which effectively captures both global shared patterns and node-specific characteristics.

{\textbf{\textit{(5)}} \textbf{Effectiveness of the parallel design of MSMA and PRL.}
We compare our parallel architecture with two cascaded variants, \textbf{P$\rightarrow$M} and \textbf{M$\rightarrow$P}. Both cascaded designs lead to performance degradation across datasets, while our parallel architecture consistently achieves better results. This suggests that MSMA and PRL provide complementary information from distinct perspectives: MSMA captures dynamic spatial dependencies, whereas PRL models long-term shared patterns and node-specific variations. Learning them independently and fusing their representations therefore better preserves their complementary benefits than sequentially transforming one representation through the other.}

{\textbf{\textit{(6)}} \textbf{Effectiveness of shared QKV across temporal scales.}
We further investigate whether scale-specific QKV projections and representations are necessary by comparing \textbf{PerScale}, which uses independent QKV projections for each scale, and \textbf{ScaleRepr}, which additionally employs scale-specific representations. Both variants show comparable or inferior performance to our shared-QKV design across datasets. This indicates that shared QKV projections are sufficient because all scales operate on the original-scale representations, while temporal downsampling and scale-specific masks already provide the necessary multi-scale structural differences.}

\begin{table}[htbp]

\centering
\setlength{\tabcolsep}{1pt}
\caption{Compatibility analysis of PRL module across different baseline models on PEMS03, PEMS04 and PEMS08.}
\renewcommand{\arraystretch}{1.8}
\resizebox{\linewidth}{!}{%
\begin{tabular}{c c ccc ccc ccc ccc}
\toprule
& \textbf{Models}
& \multicolumn{3}{c}{\textbf{RAST}}
& \multicolumn{3}{c}{\textbf{STDN}}
& \multicolumn{3}{c}{\textbf{GWNet}}
& \multicolumn{3}{c}{\textbf{STID}}
\\

\midrule
& \textbf{Metrics}
& \ding{55} & \ding{51} & \textbf{Impr(\%)}
& \ding{55} & \ding{51} & \textbf{Impr(\%)}
& \ding{55} & \ding{51} & \textbf{Impr(\%)}
& \ding{55} & \ding{51} & \textbf{Impr(\%)}
\\

\midrule
\multirow{2}{*}{\textbf{PEMS03}}

& MAE
& 15.94 & 15.82 & \cellcolor{gray!20} \textbf{$\uparrow$ 0.75}
& 15.34 & 15.34 & \cellcolor{gray!20} -
& 14.55 & 14.43 & \cellcolor{gray!20} \textbf{$\uparrow$ 0.82}
& 15.41 & 15.27 & \cellcolor{gray!20} \textbf{$\uparrow$ 0.91}
\\

& RMSE
& 27.17 & 27.13 & \cellcolor{gray!20} \textbf{$\uparrow$ 0.15}
& 27.12 & 26.84 & \cellcolor{gray!20} \textbf{$\uparrow$ 1.03}
& 25.66 & 25.01 & \cellcolor{gray!20} \textbf{$\uparrow$ 2.53}
& 27.42 & 25.10 & \cellcolor{gray!20} \textbf{$\uparrow$ 8.46}
\\

\midrule
\multirow{2}{*}{\textbf{PEMS04}}

& MAE
& 19.48 & 19.14 & \cellcolor{gray!20} \textbf{$\uparrow$ 1.75}
& 18.79 & 18.79 & \cellcolor{gray!20} -
& 18.80 & 18.73 & \cellcolor{gray!20} \textbf{$\uparrow$ 0.37}
& 18.47 & 18.36 & \cellcolor{gray!20} \textbf{$\uparrow$ 0.60}
\\

& RMSE
& 31.08 & 30.61 & \cellcolor{gray!20} \textbf{$\uparrow$ 1.51}
& 34.13 & 32.13 & \cellcolor{gray!20} \textbf{$\uparrow$ 5.86}
& 30.33 & 30.19 & \cellcolor{gray!20} \textbf{$\uparrow$ 0.46}
& 30.02 & 29.91 & \cellcolor{gray!20} \textbf{$\uparrow$ 0.37}
\\

\midrule
\multirow{2}{*}{\textbf{PEMS08}}

& MAE
& 15.14 & 14.90 & \cellcolor{gray!20} \textbf{$\uparrow$ 1.59}
& 14.41 & 14.27 & \cellcolor{gray!20} \textbf{$\uparrow$ 0.97}
& 14.59 & 14.53 & \cellcolor{gray!20} \textbf{$\uparrow$ 0.41}
& 14.22 & 14.13 & \cellcolor{gray!20} \textbf{$\uparrow$ 0.63}
\\

& RMSE
& 24.16 & 23.90 & \cellcolor{gray!20} \textbf{$\uparrow$ 1.08}
& 23.89 & 23.64 & \cellcolor{gray!20} \textbf{$\uparrow$ 1.05}
& 23.57 & 23.40 & \cellcolor{gray!20} \textbf{$\uparrow$ 0.72}
& 23.33 & 23.28 & \cellcolor{gray!20} \textbf{$\uparrow$ 0.21}
\\

\bottomrule
\end{tabular}
}
\vspace{-3mm}
\label{tab:PRL_compatibility}
\end{table}

\subsection{Prototype Residual Compatibility Analysis}
\label{sub:PRL_compatibilty}
We validate PRL's plug-and-play capability, as it addresses the universal challenge of disentangling global patterns from node-specific variations across diverse architectures.
To demonstrate PRL's generalizability, we integrate it into four representative baselines: RAST, STDN, GWNet, and STID. 
Specifically, we insert the PRL module between the hidden layers and output layers of each baseline model. The output of the baselines' hidden layers is fed into the PRL module to generate prototype-based residual features. 
These features are then concatenated with the original hidden representations before being passed to the output layer. 
This design enhances the existing models by injecting global shared patterns and node-specific deviations without modifying their core architectures.

Table~\ref{tab:PRL_compatibility} presents the performance comparison without (\ding{55}) and with (\ding{51}) the PRL module across three datasets. The results demonstrate that integrating PRL consistently improves performance across baseline models in almost all cases.
On PEMS03, GWNet and STID achieve RMSE reductions of 2.53\% and 8.46\%, while RAST on PEMS04 gains 1.75\% in MAE and 1.51\% in RMSE. On PEMS08, all four models improve consistently, with MAE and RMSE gains ranging from 0.41\% to 1.59\% and 0.21\% to 1.08\%, respectively.

These results validate two key insights: \textbf{\textit{(1)}} The PRL module can serve as an effective, architecture-agnostic component that can be seamlessly integrated into different spatial-temporal forecasting architectures as a plug-and-play component. \textbf{\textit{(2)}} The improvements are more pronounced in RMSE than MAE, indicating that PRL reduces large forecasting errors by leveraging shared global knowledge while preserving node-specific characteristics. 
Overall, PRL provides a generalizable mechanism for enhancing spatial-temporal forecasting models.

\begin{table}[!t]
\centering
\setlength{\tabcolsep}{1pt}
\caption{Analysis of residual construction strategies for PRL on PEMS03, PEMS04 and PEMS08. 
}
\renewcommand{\arraystretch}{1.4}
\resizebox{\linewidth}{!}{%
\begin{tabular}{c ccc ccc ccc}
\toprule
\textbf{Method}
& \multicolumn{3}{c}{\textbf{PEMS03}}
& \multicolumn{3}{c}{\textbf{PEMS04}}
& \multicolumn{3}{c}{\textbf{PEMS08}}\\
\midrule
\textbf{Metrics}
& MAE & RMSE & MAPE(\%)
& MAE & RMSE & MAPE(\%)
& MAE & RMSE & MAPE(\%) \\
\midrule

\textbf{Global Mean}
& 14.75	& 26.76	& 15.76
& 18.40 & 30.50 & 12.76
& 14.14 & 23.23 & 9.28
\\

\textbf{K-means Cluster}
& 14.74	& \underline{25.57}	& 15.55
& \underline{18.29} & \underline{30.20} & \underline{12.57}
& 14.10 & \underline{23.19} & 9.22
\\

\textbf{Soft Cluster}
& \underline{14.68}	& 25.80	& \underline{15.33}
& \underline{18.29} & 30.21 & 12.68
& \underline{14.06} & 23.23 & \underline{9.21}
\\

\midrule

\textbf{\name}
& \textbf{14.65} & \textbf{24.68} & \textbf{15.17}
& \textbf{18.20} & \textbf{29.89} & \textbf{12.43}
& \textbf{14.02} & \textbf{23.10} & \textbf{9.15}
\\

\bottomrule
\end{tabular}
}
\vspace{-3mm}
\label{tab:resideual construction strategies}
\end{table}

\begin{figure}[t]

    \centering
    \subfloat[Power Spectral Density (PSD) analysis.]{%
        \includegraphics[width=1.0\linewidth]{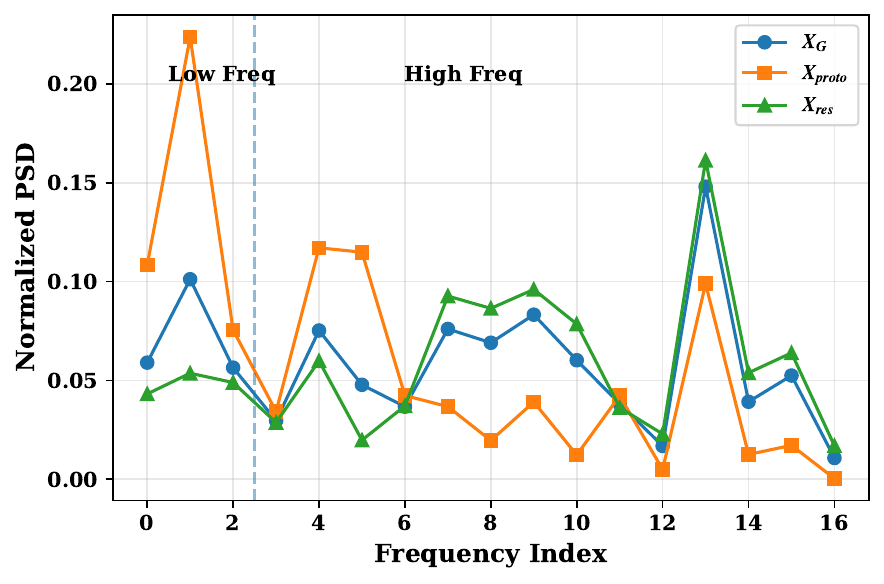}
    }
    
    \vspace{0mm}
    
    \subfloat[Low/high-frequency energy comparison.]{%
        \includegraphics[width=1.0\linewidth]{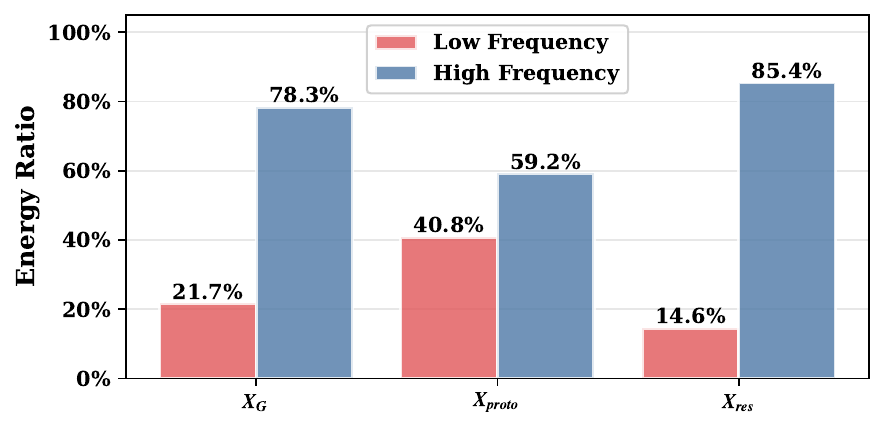}
    }
    
    \caption{Power Spectral Density (PSD) and Low/High-Frequency Energy comparison of global, prototype, and residual representations on PEMS08.}
    \label{fig:PSD_LH}
    \vspace{-6mm}
\end{figure}

\subsection{Prototype Residual Effectiveness Analysis}
\label{sub:PRL_effect}
To verify the effectiveness of the proposed Prototype Residual Learning (PRL), we compare it against three representative decomposition strategies: \textbf{\textit{(1)}} \textit{\textbf{Global Mean subtraction}}, where a shared global average representation is removed; \textbf{\textit{(2)}} \textit{\textbf{K-means Cluster Center subtraction}}, where each node is assigned to a cluster and the corresponding cluster centroid is subtracted; and \textbf{\textit{(3)}} \textit{\textbf{Soft Cluster Subtraction}}, where soft assignment weights are first computed for each node, and cluster centers are then obtained as weighted averages of node representations within the current batch.
As shown in Table~\ref{tab:resideual construction strategies}, PRL consistently achieves the best forecasting performance across all datasets. 
These results indicate that simply removing a global or cluster-level trend is insufficient for capturing the complex shared structures among traffic nodes. In contrast, the learnable prototypes provide a more expressive representation of common spatial-temporal patterns while preserving discriminative node-specific information through the residual.

{To further understand the behavior of PRL, we analyze the frequency characteristics of the decomposed representations using FFT-based Power Spectral Density (PSD). Figure~\ref{fig:PSD_LH} (a) compares the spectra of the original graph representation $\mathcal{X}_G$, the prototype representation $\mathcal{X}_{proto}$, and the residual representation $\mathcal{X}_{res}$. 
$\mathcal{X}_{proto}$ exhibits a substantially stronger concentration of low-frequency energy. Compared with $X_G$, the dominant low-frequency peak increases from 0.101 to 0.224, corresponding to an increase of approximately 121\%. Meanwhile, the residual representation suppresses low-frequency components and preserves relatively stronger responses in higher-frequency regions. For example, frequencies around bins 7--10 show noticeably larger amplitudes in $\mathcal{X}_{res}$ than in $\mathcal{X}_{proto}$, indicating that node-specific fluctuations are mainly retained in the residual representation. }

{Quantitatively, Figure~\ref{fig:PSD_LH} (b) shows that $\mathcal{X}_{proto}$ concentrates approximately 40.8\% of its spectral energy in the low-frequency band, while $\mathcal{X}_{res}$ contains only about 14.6\%. Conversely, $\mathcal{X}_{res}$ allocates substantially more energy to higher-frequency components, resulting in a low-frequency energy gap of approximately 26.2\%. This clear spectral redistribution demonstrates that PRL effectively separates globally shared trends from localized node-specific variations, rather than merely removing information from the original representation.}

\begin{figure}[t]
    \centering
    \setlength{\tabcolsep}{0mm}
    \begin{tabular}{cccc}
        \subfloat[PEMS03 $R$]{\includegraphics[width=0.5\linewidth]{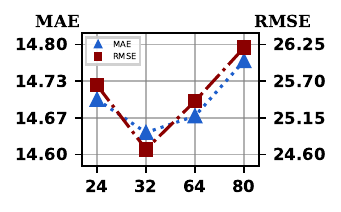}} &
        \subfloat[PEMS03 $F$]{\includegraphics[width=0.5\linewidth]{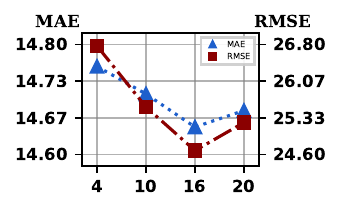}} \\[-1mm]
        \subfloat[PEMS03 $\tau$]{\includegraphics[width=0.5\linewidth]{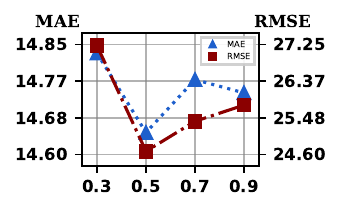}} &
        \subfloat[PEMS03 $\mu$]{\includegraphics[width=0.5\linewidth]{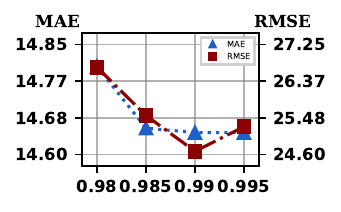}} 
        \\[-1mm]
        \vspace{0mm}
        \subfloat[PEMS04 $R$]{\includegraphics[width=0.5\linewidth]{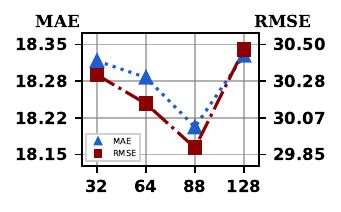}} &
        \subfloat[PEMS04 $F$]{\includegraphics[width=0.5\linewidth]{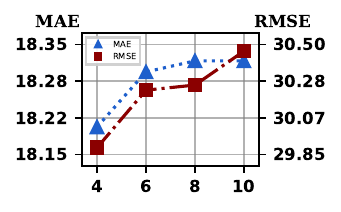}} \\[-1mm]
        \vspace{0mm}
        \subfloat[PEMS04 $\tau$]{\includegraphics[width=0.5\linewidth]{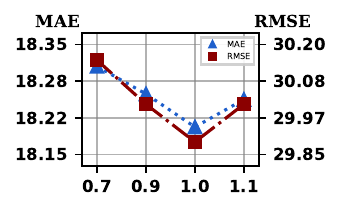}} &
        \subfloat[PEMS04 $\mu$]{\includegraphics[width=0.5\linewidth]{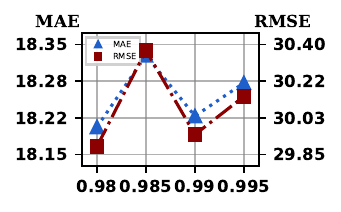}} 
        \\[-1mm]
    \end{tabular}
    \caption{Hyper-parameter analysis.}
    \label{fig:hyperparameters-pem03-pems04}
    \vspace{-6mm}
\end{figure}

\subsection{Hyper-Parameter Analysis}
\label{sub:hyper}
In this subsection, we analyze the sensitivity of our model to key hyperparameters. We evaluate the number of prototypes $R \in \{24, 32, 64, 80\}$ (PEMS03) and $R \in \{32, 64, 88, 128\}$ (PEMS04), the feature dimension $F \in \{4, 10, 16, 20\}$ (PEMS03) and $F \in \{4, 6, 8, 10\}$ (PEMS04), the temperature parameter $\tau \in \{0.3, 0.5, 0.7, 0.9\}$ (PEMS03) and $\tau \in \{0.7, 0.9, 1.0, 1.1\}$ (PEMS04), and the momentum coefficient $\mu \in \{0.98, 0.985, 0.99, 0.995\}$ (PEMS03 and PEMS04) for prototype updating. 
From the results shown in Figure~\ref{fig:hyperparameters-pem03-pems04}, several conclusions can be made:
\textbf{\textit{(1)}} As shown in Figure~\ref{fig:hyperparameters-pem03-pems04} (a) and (e), {the performance improves first and then degrades as $R$ increases, with optimal values of $R=32$ (PEMS03) and $R=88$ (PEMS04). This suggests that too few prototypes cannot fully capture diverse global patterns, while too many may introduce redundancy. Although PEMS03 has more nodes and a longer time span, it requires fewer prototypes. This indicates that PEMS03 may exhibit more homogeneous spatial patterns, while PEMS04 exhibits greater heterogeneity requiring more prototypes to capture diverse behaviors.
\textbf{\textit{(2)}} As shown in Figure~\ref{fig:hyperparameters-pem03-pems04} (b) and (f), PEMS03 achieves optimal performance at $F=16$, while PEMS04 performs best with $F=4$. This indicates that PEMS03 benefits from higher-dimensional feature representations to capture richer temporal dynamics, while the smaller one requires lower-dimensional features to avoid overfitting.
\textbf{\textit{(3)}} As shown in Figure~\ref{fig:hyperparameters-pem03-pems04} (c) and (g), the temperature parameter $\tau$ controls the sharpness of prototype assignments, with optimal values of $\tau=0.5$ for PEMS03 and $\tau=1.0$ for PEMS04. 
The lower optimal temperature for PEMS03 suggests its homogeneous patterns benefit from sharper assignments, while PEMS04's higher temperature indicates its heterogeneous patterns require softer assignments to flexibly associate with multiple prototypes.}
{\textbf{\textit{(4)}} As shown in Figure~\ref{fig:hyperparameters-pem03-pems04} (d) and (h), the momentum coefficient $\mu$ controls the update rate of prototypes. A smaller $\mu$ makes prototypes more responsive to newly observed traffic patterns, whereas a larger $\mu$ emphasizes historical information and yields more stable updates. The best performance is achieved at $\mu=0.995$ on PEMS03 and $\mu=0.98$ on PEMS04. This observation is consistent with the analysis of $R$ in \textbf{\textit{(1)}}: PEMS03 appears to exhibit more homogeneous spatial patterns, allowing prototypes to evolve slowly while maintaining representative global semantics, whereas the greater heterogeneity of PEMS04 benefits from faster prototype adaptation to capture more diverse traffic behaviors.}

\subsection{Computational Efficiency Study}
\label{sub:computation}
Our framework maintains computational efficiency through two designs. MSMA generates lightweight binary masks via 1D convolutions and shares QKV projections across scales, avoiding repeated attention computation. PRL employs a compact prototype bank with $R \ll N$ prototypes, enabling global pattern modeling without pairwise node-wise interactions. We compare training time, inference time, model parameters, and memory usage with five baselines on PEMS08 (170 nodes) and PEMS07 (883 nodes) in Table~\ref{tab:computational}.

Our model achieves the fastest training and inference speed on PEMS08, requiring only 12.67s / 1.69s, and maintains competitive efficiency on PEMS07 with 219.18s / 21.38s. Notably, compared with complex models such as STDN and STAEFormer, which require significantly longer training time, our method demonstrates substantial efficiency gains.

In terms of model complexity, our architecture is compact (279K parameters on PEMS08 and 301K on PEMS07), comparable to lightweight models like RAST, and significantly smaller than STDN and STAEFormer. This compactness is mainly attributed to shared QKV projections in MSMA and the compact prototype bank in PRL, which avoid redundant parameterization and pairwise interactions. 

Despite modeling multi-scale spatial dependencies and maintaining a prototype bank, \name requires only 2,806 MB and 9,030 MB of memory on PEMS08 and PEMS07, respectively, substantially less than STAEFormer and $D^2$STGNN.

Overall, the results demonstrate that MSMA and PRL jointly reduce both temporal modeling and global interaction costs, enabling efficient yet expressive spatial-temporal forecasting across datasets of different scales. This leads to an excellent balance between predictive accuracy and computational efficiency, making the proposed framework well-suited for deployment in resource-constrained environments while maintaining state-of-the-art forecasting performance.

\begin{table}[!t]
\centering
\setlength{\tabcolsep}{1pt}
\caption{Efficiency comparison on PEMS08 and PEMS07.}
\renewcommand{\arraystretch}{1.0}
\resizebox{\linewidth}{!}{%
\begin{tabular}{c cccc cccc}
\toprule
\textbf{Dataset}
& \multicolumn{4}{c}{\textbf{PEMS08}}
& \multicolumn{4}{c}{\textbf{PEMS07}}\\
\midrule
\textbf{Metrics}
& \makecell{\textbf{Train} \\ \textbf{Time (s)}}
& \makecell{\textbf{Infer} \\ \textbf{Time (s)}}
& \makecell{\textbf{Params} \\ \textbf{(K)}}
& \makecell{\textbf{Mem Usage} \\ \textbf{(MB)}}

& \makecell{\textbf{Train} \\ \textbf{Time (s)}}
& \makecell{\textbf{Infer} \\ \textbf{Time (s)}}
& \makecell{\textbf{Params} \\ \textbf{(K)}}
& \makecell{\textbf{Mem Usage} \\ \textbf{(MB)}}

\\
\midrule
\textbf{$D^2$STGNN}
& 34.94 & 4.55 & 391 & 8296 & 672.45 & 87.32 & 405 & 15613
\\
\textbf{STAEFormer}
& 35.11 & 4.56 & 1223 & 8018 & 458.81 & 55.29 & 1908 & 22919
\\
\textbf{STWave}
& 31.47 & 5.25 & 882 & 5660 & 247.83 & 39.87 & 882 & \underline{8539}
\\
\textbf{STDN}
& 39.93 & 4.79 & 5876 & 9028 & 293.39 & 31.05 & 4489 & 9213
\\
\textbf{RAST}
& \underline{15.58} & \underline{4.79} & \textbf{265} & \textbf{1244} & \textbf{76.78} & \textbf{20.78} & \underline{311} & \textbf{3253}
\\
\midrule
\textbf{\name}
& \textbf{12.67} & \textbf{1.69} & \underline{279} & \underline{2806} & \underline{219.18} & \underline{21.38} & \textbf{301} & 9030
\\

\bottomrule
\end{tabular}
}
\vspace{-3mm}
\label{tab:computational}
\end{table}

\subsection{MSMA Generalization Analysis}
\label{sub:MSMA gene}
To validate the effectiveness of the proposed binary mask sampling mechanism, we compare Binary Mask Sampling with two alternative dependency modeling strategies: \textbf{\textit{(1)}} {\textbf{Continuous Similarity Weighting}}, which directly uses continuous similarity scores as spatial weights, and \textbf{\textit{(2)}} {\textbf{Adaptive Attention Weighting}}, which models dependencies via a dense learnable attention matrix. The results are reported in Table~\ref{tab:gumbel}.

Binary masks are designed by filtering weak and noisy correlations in continuous similarity weights, enforcing sparsity to focus on the most informative spatial dependencies. Compared with Continuous Similarity Weighting, binary mask sampling explicitly removes less relevant interactions, thereby reducing noise propagation during message passing. Compared with Adaptive Attention Weighting, which models dense pairwise relations among all nodes, the sampled masks induce sparse and adaptive graph structures that better capture critical spatial dependencies while avoiding redundant interactions.
As shown in Table~\ref{tab:gumbel}, the proposed method consistently achieves the best overall performance across datasets. These results suggest that explicit structure selection is more effective than dense relation weighting for spatial dependency modeling. 

\begin{figure}[t]
    \centering
    \subfloat[PEMS04]{\includegraphics[width=0.5\linewidth]{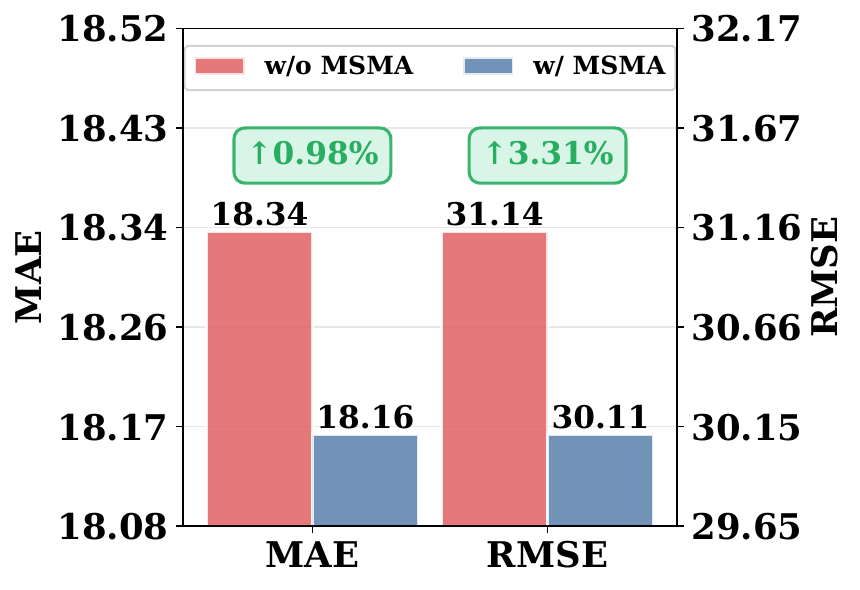}}
    \hfill
    \subfloat[PEMS08]{\includegraphics[width=0.5\linewidth]{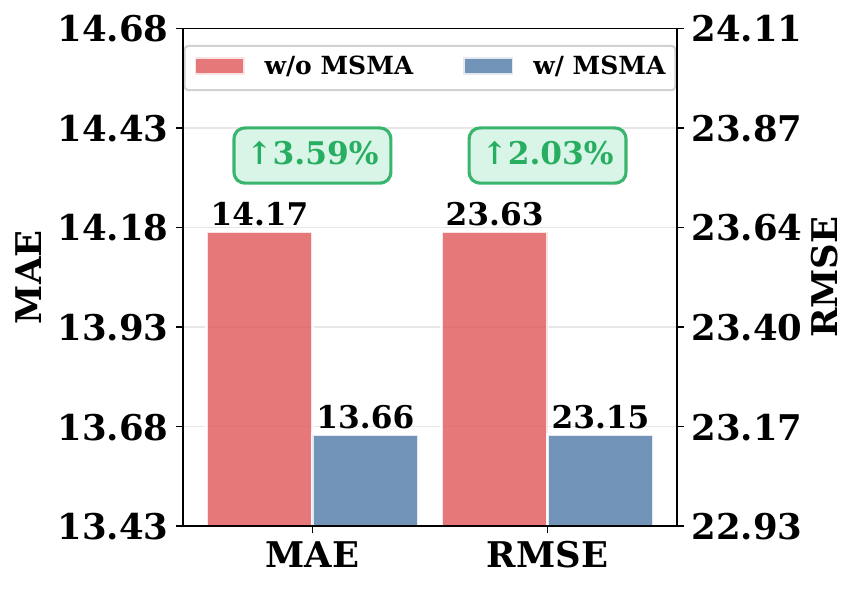}}
    \caption{Compatibility analysis of MSMA with STAEFormer on PEMS04 and PEMS08.}
    \label{fig: msma_compatibility_analysis}
    \vspace{-3mm}
\end{figure}

To evaluate the compatibility of the proposed MSMA module, we integrate it into the STAEFormer~\cite{liu2023spatio} backbone and compare the performance with and without MSMA on the PEMS04 and PEMS08 datasets. As shown in Figure~\ref{fig: msma_compatibility_analysis}, the inclusion of MSMA consistently improves forecasting accuracy across both datasets and evaluation metrics. 
These consistent gains indicate that MSMA can be incorporated into existing spatial-temporal backbones and effectively enhance their ability to capture multi-scale dependency patterns, demonstrating compatibility and plug-and-play capability.

\begin{table}[!t]
\centering
\setlength{\tabcolsep}{1pt}
\caption{Comparative analysis of MSMA module on PEMS03, PEMS04 and PEMS08. 
}
\renewcommand{\arraystretch}{1.4}
\resizebox{\linewidth}{!}{%
\begin{tabular}{c ccc ccc ccc}
\toprule
\textbf{Method}
& \multicolumn{3}{c}{\textbf{PEMS03}}
& \multicolumn{3}{c}{\textbf{PEMS04}}
& \multicolumn{3}{c}{\textbf{PEMS08}}\\
\midrule
\textbf{Metrics}
& MAE & RMSE & MAPE(\%)
& MAE & RMSE & MAPE(\%)
& MAE & RMSE & MAPE(\%) \\
\midrule

\textbf{Continuous Similarity}
& 14.81	& 27.19	& 15.67
& \underline{18.28} & \underline{30.16} & 12.55
& \textbf{14.02} & 23.15 & 9.33
\\

\textbf{Adaptive Attention}
& \underline{14.72} & \underline{25.83} & \underline{15.34}
& 18.34 & 30.31 & \underline{12.51}
& \underline{14.08} & \underline{23.12} & \underline{9.21}
\\

\midrule

\textbf{\name}
& \textbf{14.65} & \textbf{24.68} & \textbf{15.17}
& \textbf{18.20} & \textbf{29.89} & \textbf{12.43}
& \textbf{14.02} & \textbf{23.10} & \textbf{9.15}
\\

\bottomrule
\end{tabular}
}
\label{tab:gumbel}
\end{table}

\begin{figure}[t]
    \centering
    \includegraphics[width=\linewidth]{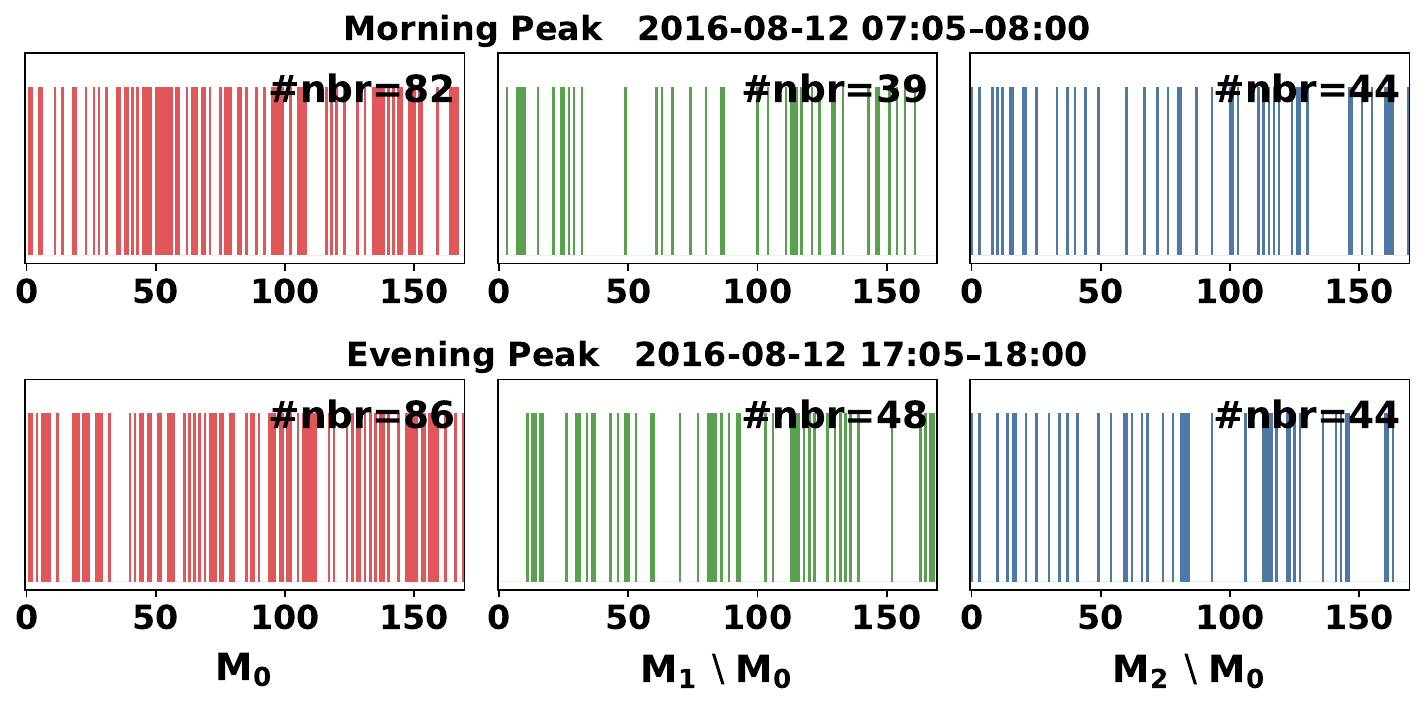} 
    \caption{Case study of multi-scale mask for node 91.}
    \label{fig:mask_union}
    \vspace{-3mm}
\end{figure}

\subsection{Case Study of Multi-Scale Mask Complementarity Analysis}
\label{sub:Mask_com}
To examine how the complementary spatial patterns identified by different temporal scales manifest at the sensor level, we conduct a case study using morning- and evening-peak traffic windows, as shown in Fig.~\ref{fig:mask_union}. For a representative sensor, we visualize the neighbors activated by $\mathbf{M}^{(0)}$, as well as those additionally activated by $\mathbf{M}^{(1)}$ and $\mathbf{M}^{(2)}$ after excluding the neighbors identified by $\mathbf{M}^{(0)}$. Across scales, the masks identify both overlapping and scale-specific neighbors, indicating complementary spatial coverage. Across peak periods, the composition and number of identified neighbors vary between morning and evening, suggesting that the learned masks adapt to different traffic conditions. Overall, this case study illustrates that MSMA generates multi-scale spatial masks that capture complementary spatial dependencies and adapt to temporal variations in traffic patterns.

\begin{table}[htbp]
\centering
\setlength{\tabcolsep}{1pt}
\caption{Scalability evaluation on GBA dataset. 
}
\renewcommand{\arraystretch}{1.0}
\resizebox{\linewidth}{!}{%
\begin{tabular}{c ccccccc}
\toprule
\textbf{Metrics}
& MAE & RMSE & MAPE(\%)
& \makecell{Train Time (s)}
& \makecell{Infer Time (s)}
& \makecell{Params (K)}
& Batch Size
\\
\midrule

STAEformer
& 21.30	& 34.56	& 17.63
& 2226.32 & 247.46 & 3318 & 8

\\

DGCRN
& \underline{20.53} & 34.40 & 16.79
& 2157.33 & 618.62 & \underline{371} & 32

\\

STWave
& 20.56 & \underline{33.98} & \textbf{15.14}
& 551.74 & \underline{91.33} & 882 & 64

\\

RAST
& 20.64 & 34.47 & 16.63
& \textbf{323.59} & 101.54 & 1491 & 64
\\

\midrule

\name
& \textbf{20.05} & \textbf{33.69} & \underline{15.97}
& \underline{350.46} & \textbf{50.77} & \textbf{325} & 64

\\

\bottomrule
\end{tabular}
}
\label{tab:gba}
\vspace{-3mm}
\end{table}

    
    
    

\subsection{Scalability Evaluation on Larger Graphs}
\label{sub:Large}
To evaluate the scalability of \name on larger graphs, we conduct additional experiments on the GBA dataset with 2,352 nodes. All experiments are conducted on an NVIDIA A800 GPU under the same setting. 
As shown in Table~\ref{tab:gba}, \name achieves the best MAE and RMSE among all compared methods, while obtaining competitive MAPE. In terms of computational efficiency, \name achieves the lowest inference time of 50.77s, reducing inference time by approximately 44\% compared with STWave, the second-fastest method. Its training time is 350.46s, close to that of RAST (323.59s) and substantially shorter than those of STAEformer and DGCRN. Moreover, \name uses only 325K parameters, fewer than all compared methods except DGCRN, while supporting a batch size of 64. These results indicate that \name achieves a favorable balance between forecasting accuracy and computational efficiency, making it suitable for larger-scale traffic networks.

We note that the spatial attention module has quadratic computational complexity with respect to the number of nodes, i.e., \(O(N^2)\). To reduce the memory overhead associated with large-scale attention computation, without modifying the architecture or formulation of \name, we employ the memory-efficient kernel~\cite{rabe2021self} provided by PyTorch's scaled dot-product attention (SDPA). This implementation avoids explicitly materializing the full \(N\times N\) attention matrix, reducing intermediate memory requirements during attention computation. This implementation helps maintain practical computational efficiency when modeling spatial dependencies over large graphs.

%% file: sections/3_RelatedWorks.tex
\section{Related Work}

The evolution of traffic forecasting has been driven by increasingly expressive models for capturing spatial-temporal dependencies~\cite{zheng2020gman, dong2024heterogeneity, ma2025less, gao2025different}. Early works such as GWNet~\cite{wu2019graph} and AGCRN~\cite{bai2020adaptive} integrated adaptive graph learning with recurrent architectures, establishing the foundation for spatial-temporal modeling. Recent advances have pursued three complementary directions. First, dynamic graph learning methods recognize that spatial correlations evolve over time. DGCRN~\cite{li2023dynamic} generates time-varying graphs through hyper-networks, while Transformer-based approaches like PDFormer~\cite{jiang2023pdformer} and STAEformer~\cite{liu2023spatio} capture propagation delays and intrinsic temporal-scale relations through attention mechanisms.
Second, signal decomposition frameworks disentangle complex traffic patterns into interpretable components. $D^2$STGNN~\cite{shao2022decoupled} separates diffusion from inherent signals, STDN~\cite{cao2025spatiotemporal} isolates trend-cyclical and seasonal patterns, while STWave~\cite{fang2023spatio} decomposes stable trends from high-frequency fluctuations. Third, efficient architectures address computational scalability. STID~\cite{shao2022spatial} demonstrates that simple MLPs with spatial-temporal identities achieve competitive accuracy, BigST~\cite{han2024bigst} scales to large networks via linearized convolutions, LightST~\cite{zhang2025efficient} employs knowledge distillation, and RAST~\cite{ruan2025rast} enhances forecasting through retrieval mechanisms.
However, existing methods rarely jointly address both aspects: modeling scale-dependent spatial structures and separating globally shared patterns from node-specific residual deviations.

%% file: sections/6_Conclusion.tex
\section{Conclusion}
In this paper, we address two fundamental challenges in spatial-temporal forecasting: granularity-dependent spatial correlations and signal entanglement between globally shared patterns and node-specific variations. We propose \name, a unified framework synergizing multi-scale structural modeling with prototype-based semantic decomposition. Our Multi-Scale Masked Attention (MSMA) adaptively captures scale-specific spatial dependencies by generating granularity-aware masks from multi-resolution temporal features, distinguishing micro-scale physical neighbors from macro-scale functional correlations. Complementarily, our Prototype Residual Learning (PRL) maintains a momentum-updated prototype bank to encode shared patterns, then explicitly extracts node-specific residuals through subtraction, preventing over-smoothing while preserving heterogeneity. Extensive experiments on five datasets demonstrate that \name achieves state-of-the-art performance, validating the effectiveness of jointly modeling granularity-dependent structures and prototype-based residual decomposition.

\section{Acknowledgments}
Zijian Zhang is supported by Key R\&D Project Funding of Jilin Province (No. 20260205050GH), the Open Funding Programs of State Key Laboratory of AI Safety, and China Postdoctoral Science Foundation (No. 2025M771587). 
Hao Miao is supported by SCRI, The Hong Kong Polytechnic University (No. Q-CDDG). 
Juncheng Hu is supported by the National Key Research and Development Program (No. 2024YFB3310200).
Irwin King is supported by the Research Grants Council of the Hong Kong Special Administrative Region, China (CUHK 2300246, RGC C1043-24G) and (CUHK 14203425, RGC GRF 2151317).

\section{AI-generated content acknowledgments}
Portions of this manuscript were refined using language assistance tools to improve grammar, clarity, and readability. These tools were used solely for language refinement. The authors carefully reviewed and edited all AI-assisted content and take full responsibility for the final manuscript, including its conceptualization, experiments, and conclusions.